\documentclass[10pt,twocolumn,letterpaper]{article}

\usepackage{wacv}
\makeatletter
\@namedef{ver@everyshi.sty}{}
\makeatother
\usepackage{tikz}
\usepackage{times}
\usepackage{epsfig}
\usepackage{graphicx}
\usepackage{amsmath}
\usepackage{amssymb}
\usepackage{booktabs}
\usepackage{graphicx}
\usepackage{tabularx}
\usepackage{caption}
\usepackage{adjustbox}
\usepackage{ifsym}
\usepackage{enumitem}
\usepackage{indentfirst}
\usepackage{float}
\usepackage{tikz}
\usepackage{comment}
\usepackage{amsmath,amssymb} 
\usepackage{mwe}
\usepackage{subcaption}
\usepackage{color}
\usepackage{bbold}
\newcolumntype{K}[1]{>{\centering\arraybackslash}p{#1}}
\usepackage{multirow, makecell}
\usepackage{authblk}

\def\wacvPaperID{1034} 

\wacvfinalcopy 

\ifwacvfinal
\usepackage[breaklinks=true,bookmarks=false]{hyperref}
\else
\usepackage[pagebackref=true,breaklinks=true,colorlinks,bookmarks=false]{hyperref}
\fi

\pdfoutput=1
\begin{document}

\title{Class-Level Confidence Based 3D Semi-Supervised Learning}

\author[1]{Zhimin Chen}
\author[2]{Longlong Jing}
\author[2]{Liang Yang}
\author[3]{Yingwei Li}
\author[1]{Bing Li}
\affil[1]{Clemson University} 
\affil[2]{The City University of New York}
\affil[3]{Johns Hopkins University}
\affil[ ]{\textit {\tt\small \{zhiminc,bli4\}@clemson.edu, ljing@gradcenter.cuny.edu, lyang1@ccny.cuny.edu, yingwei.li@jhu.edu}}



\maketitle
\thispagestyle{empty}
\begin{abstract}
Recent state-of-the-art method FlexMatch firstly demonstrated that correctly estimating learning status is crucial for semi-supervised learning (SSL). However, the estimation method proposed by FlexMatch does not take into account imbalanced data, which is the common case for 3D semi-supervised learning. To address this problem, we practically demonstrate that unlabeled data class-level confidence can represent the learning status in the 3D imbalanced dataset. Based on this finding, we present a novel class-level confidence based 3D SSL method. Firstly, a dynamic thresholding strategy is proposed to utilize more unlabeled data, especially for low learning status classes. Then, a re-sampling strategy is designed to avoid biasing toward high learning status classes, which dynamically changes the sampling probability of each class. To show the effectiveness of our method in 3D SSL tasks, we conduct extensive experiments on 3D SSL classification and detection tasks. Our method significantly outperforms state-of-the-art counterparts for both 3D SSL classification and detection tasks in all datasets.
\end{abstract}

\section{Introduction}

As 3D point cloud data collection and annotation is expensive and time-consuming, 3D semi-supervised learning (SSL) has attracted increasing attention in recent years and shows its superiority in utilizing the unlabeled data ~\cite{zhao2020sess,wang20213dioumatch,chen2021multimodal,sohn2020fixmatch,zhang2021flexmatch}. Most existing semi-supervised learning methods, such as Pseudo-Labeling~\cite{lee2013pseudo} and  FixMatch~\cite{sohn2020fixmatch}, employ the pseudo-labeling strategy in which the network's high confidence predictions on the unlabeled data are used as labels to further optimize the network.

The pseudo-labeling strategy based methods were widely used and achieve significant improvement in performance for many different tasks~\cite{zhao2020sess,wang20213dioumatch,chen2021multimodal,sohn2020fixmatch}. However, a  non-negligible drawback of pseudo labeling is that it relies on a manually pre-defined fixed threshold to choose high-quality pseudo labels. For each data from the unlabeled data, no matter its category, the data will be used for training only when its confidence is higher than this fixed threshold; otherwise, this data will be ignored. The common practice uses a very high threshold (0.9)~\cite{wang20213dioumatch,chen2021multimodal} in 3D tasks to keep the pseudo-labels with high quality. 
This fixed threshold ignores the different learning statuses among classes and thus left a large number of unlabeled data unused, which compromises the model performance.


\begin{figure}[ht!]
    \centering
    \begin{subfigure}[b]{0.4\textwidth}   
        \centering 
\includegraphics[width=6.5cm,height=4.5cm]{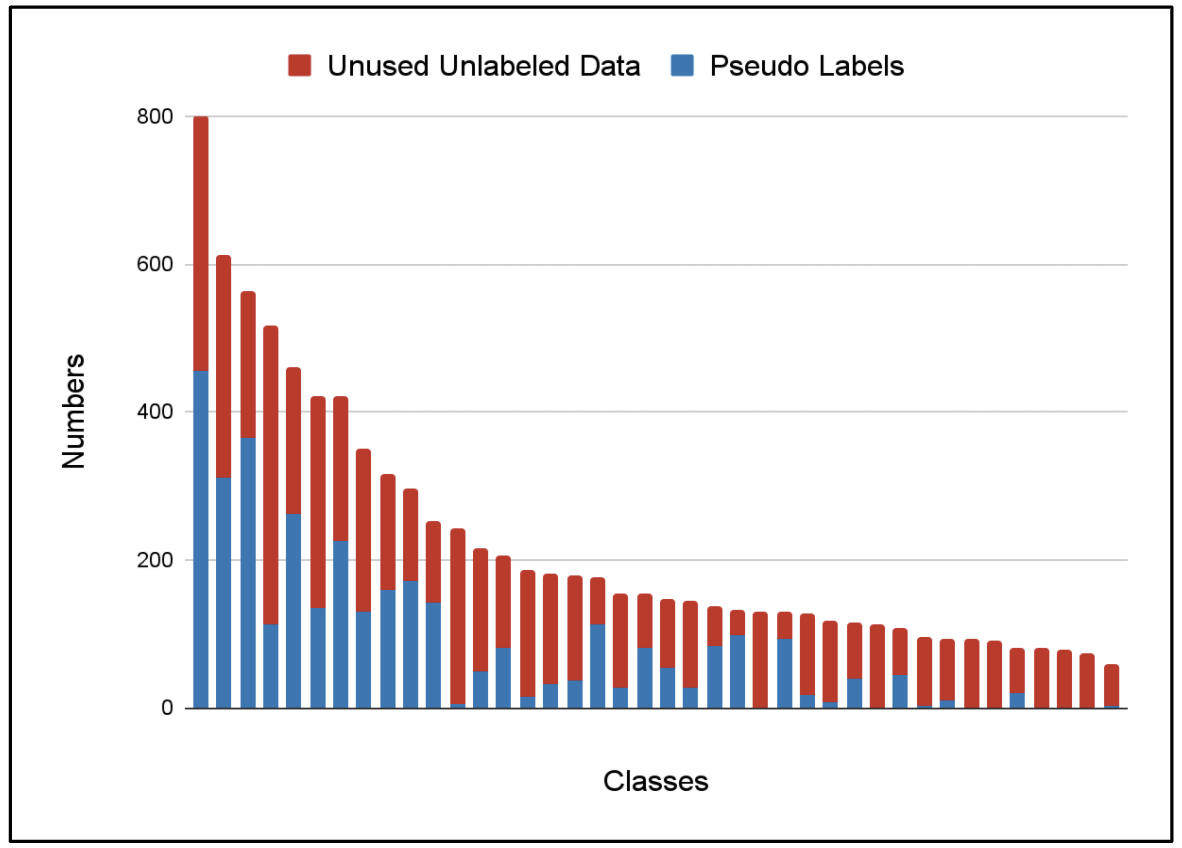}
        \caption[]%
        {{\small }}    
        \label{fig:psd_ratio}
    \end{subfigure}
    \begin{subfigure}[b]{0.4\textwidth}   
        \centering 
        \includegraphics[width=6.5cm,height=4.5cm]{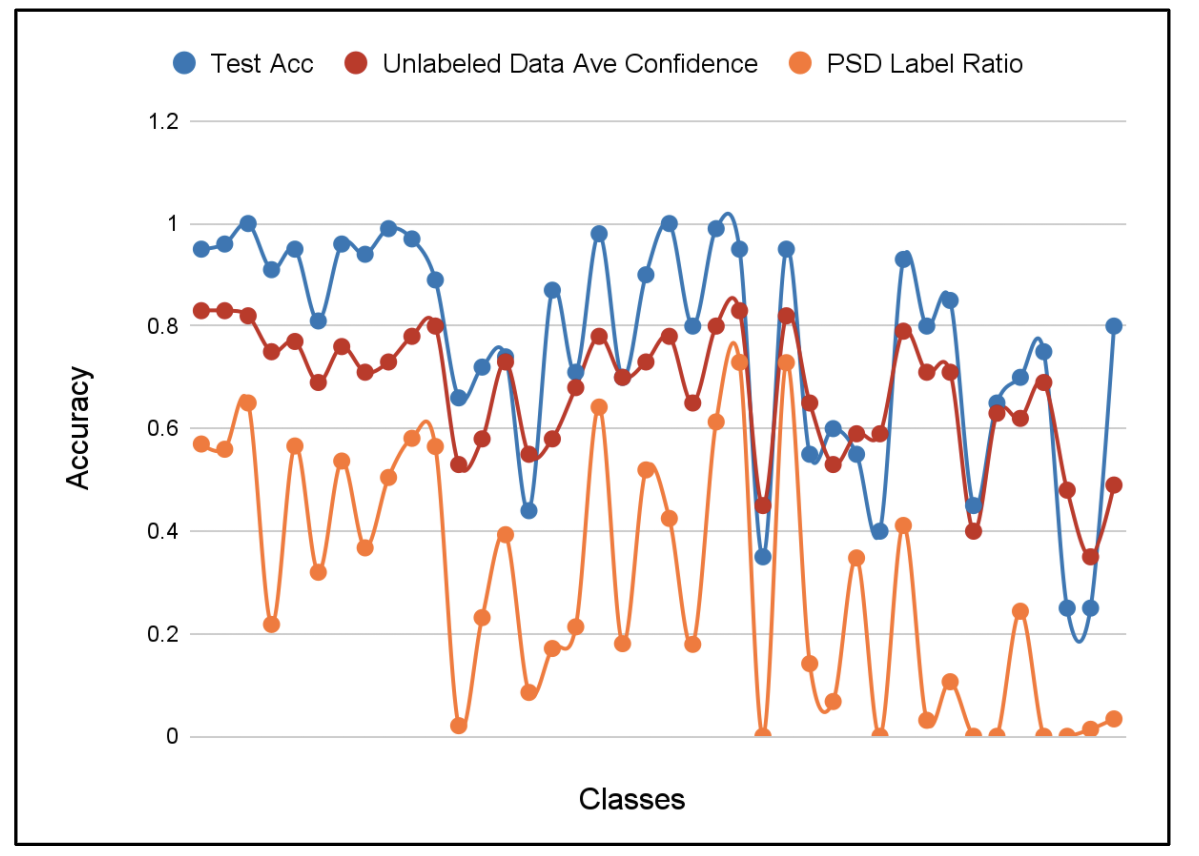}
        \caption[]%
        {{\small }}    
        \label{fig:modelnet40-confid}
    \end{subfigure}

    \caption[ ]
    {\small The results analysis of FixMatch trained in ModelNet40 dataset with 10\% labeled data. (a) The ratio of selected pseudo-labels to unlabeled data. (b) Test accuracy and the class-level confidence of unlabeled data. It is obvious to observe that (1) Only a
    small percentage of unlabeled data is used during training with high threshold setting. (2) Learning difficulty of each class classes has high variance. Some minority classes have higher accuracy than majority classes. (3) The class-Level confidence has high correlation with test accuracy of each class.} 
    \vspace{-5mm}
    \label{fig:acc-confid-relationship}
\end{figure}

 To solve this issue, an intuitive way is to estimate the learning status of each class and set dynamic threshold for each class accordingly. However, the key challenge is how to estimate the learning status. Recently SOTA method FlexMatch ~\cite{zhang2021flexmatch} leveraged curriculum learning approach to estimate learning status of each class and flexibly adjust thresholds. However, the FlexMatch does not well-define the terminology 'learning status'. In this work, we regard 'learning status' as how well the model learning for a class and can be reflected in the test accuracy. The test accuracy of each class is utilized to represent the learning status in our analysis. Furthermore, in FlexMatch, only learning difficulty is taken into account to estimate learning status but the imbalanced data condition is not included. We find that both learning difficulty and imbalanced data affect the learning status of network. For example, the network may have similar performance on high learning difficulty but majority classes and low learning difficulty but minority classes.

Hence, a more accurate and general estimation method of learning status is required for semi-supervised learning. Previous work ~\cite{lee2013pseudo,sohn2020fixmatch} utilizes instance-level confidence to represent the instance learning status even in imbalanced data and can be utilized to select well-learned pseudo labels. Therefore, it is intuitive to assume that the class-level learning status can be reflected by the class-level confidence. Furthermore, our analysis in Fig.~\ref{fig:modelnet40-confid} demonstrate that there is a high correlation between the average class-level confidence score and the test accuracy results in 3D imbalanced data. More analysis on detection task can be found in the supplementary materials. Hence, we hypothesize that the class-level confidence on the unlabeled data can be leveraged to estimate the learning status of each class in 3D imbalanced data. 


Inspired by our intuitions and the analysis above, we propose a semi-supervised learning method that can dynamically adjust the threshold based on the learning status.  The dynamic threshold is adjusted based on the learning status of each class, which allows more unlabeled data of low learning status to be utilized. Furthermore, our method can also utilize more unlabeled data at the early stage of the training process where only few data have predicted confidence larger than the fixed high thresholds. However, we find that improving numbers of selected pseudo-label with dynamic threshold cannot eliminate the learning status variance caused by data imbalance and learning difficulty variance. This makes network biased toward high learning status classes and thus overfitted. To avoid this issue, we  further propose a novel re-sampling strategy to dynamically sample the data based on the learning status. Specifically, our re-sampling strategy increases the sample probability of instances belonging to low learning status classes and decreases the sample probability of high learning status classes. With the dynamical thresholding and re-sampling, our method can utilize the unlabeled data more effectively  and balance the learning status of each class .

The goal of our method is to estimate the learning status of each class and further balance and improve the learning status.
Compared to the other 3D semi-supervised learning methods, our method estimate the learning status of each class, dynamically adjust the threshold, and re-sampling the data based on the learning status. Our method can be easily applied to different semi-supervised tasks to improve performance  even in imbalanced dataset. To demonstrate the generality of the proposed method, we evaluated our proposed method on two different tasks, including SSL 3D object recognition and SSL 3D object detection tasks. Our method outperforms the state-of-the-art methods by a large margin.

Our key contributions are summarized as follows: 
\begin{enumerate}[itemindent=1em]
    \item 
    We clarify the definition of learning status and practically demonstrate that the class-level confidence can represent the learning status of each class. Based on this finding, we propose a learning status estimation method that works well in the imbalanced 3D dataset.
    
    \item We firstly incorporate learning difficult and imbalanced data problems together based on learning status. We propose a novel 3D semi-supervised learning method to dynamically adjust the thresholds and re-sample data based on each class learning status, which balances and improve learning status and thus solves the variance of learning difficult and imbalanced data problems at the same time. 
    
    \item Our proposed method outperforms the state-of-the-art semi-supervised 3D object detection and classification methods by a large margin.
\end{enumerate}

\section{Related Work}
\textbf{Semi-Supervised Learning:} Semi-supervised learning methods have made a significant progress in recently years ~\cite{berthelot2019mixmatch,chen2020big,ghosh2021data,lee2013pseudo,miyato2018virtual,tarvainen2017mean,oliver2018realistic,xu2021dash}. Many existing SSL methods utilize pseudo-labeling~\cite{lee2013pseudo} to minimizes the entropy of the predictions on unlabeled data. The performance of pseudo labeling heavily relies on the quality of pseudo-labels. To improve the quality of pseudo-labels, state-of-the-art SSL method FixMatch~\cite{sohn2020fixmatch} and many FixMatch-liked methods usually set a fixed high confidence threshold to filter out low-confidence predictions from the strong augmented data. The high-value threshold can improve the quality of pseudo labels but ignores the learning status of each class, which not only engenders the bias toward high learning status classes but leaves a large number of unlabeled data unused. To address this issue, Dash~\cite{xu2021dash} uses cross entropy loss to obtain dynamic threshold for all classes. FlexMatch ~\cite{zhang2021flexmatch} substitutes the pre-defined threshold with flexible thresholds based on curriculum learning to consider each class learning status. However, our experiments demonstrate that the FlexMatch does not generalize well in 3D dataset. This is because it is designed for class-balanced datasets, but the 3D dataset is inherently imbalanced. Furthermore, in 3D networks, the prediction confidence of some classes cannot achieve the pre-defined high threshold and make those classes thresholds extremely small. Therefore, the learning status estimation strategy in FlexMatch is not suitable to be applied in 3D SSL tasks.  More comparison with FlexMatch can be found in supplemental materials.

 \textbf{Class-Imbalanced Semi-Supervised Learning:} 
Recently, class imbalanced semi-supervised learning has attracted increasing attention as it more accurately describes the real-world data distribution ~\cite{yang2020rethinking,hyun2020class}. Wei et al ~\cite{wei2021crest} found that the raw SSL methods usually have high recall and low precision for head classes and proposed CReST to re-sampling unlabeled data based on the labeled data numbers. Recent state-of-the-art method BiS ~\cite{he2021rethinking} deployed two different re-sampling strategies at the same time to decoupled train the model. All of those class-imbalanced SSL methods resample based on data numbers. However, we find that some minority classes may have better performance than majority classes due to their low learning difficulty. Sampling based on data numbers makes the model biased toward those low learning difficulty classes. Due to the page limitation, we add the comparison with state-of-the-art imbalanced SSL methods in supplemental materials.

\textbf{3D Semi-Supervised Object Classification:} Currently, many 3D object classification method have been proposed for 3D understanding~\cite{qi2017pointnet,qi2017pointnet++,thomas2019kpconv,wu2019pointconv,xu2018spidercnn,zhao2019pointweb,zhao2020point,xu2020weakly,xu2020weakly}. Unlike 2D features, a 3D model is complex by nature and thus hard to extract. To better extract 3D features, many works~\cite{li2019deepgcns,qi2017pointnet,qi2017pointnet++,dgcnn} have been proposed to extract features from 3D point clouds for the 3D classification tasks. However, 3D SSL classification is underexploited. Chen et al.~\cite{chen2021multimodal} finds that directly implementing 2D SSL methods like FixMatch~\cite{sohn2020fixmatch}, S4L~\cite{zhai2019s4l}, and Pseudo label~\cite{lee2013pseudo} cannot achieve comparable results in 2D and propose to utilize multi-modality information in 3D SSL classification. However, what downgrades those 2D SSL methods' performance in 3D is left to explore. 

\textbf{3D Semi-Supervised Object Detection:} According to the input data formats, current methods for 3D object detection task can be summarized into three different types: 2D projection ~\cite{li2019gs3d,simony2018complex,yang2018pixor}, voxel grid~\cite{lang2019pointpillars,ren2016three,song2016deep}, and point cloud~\cite{lahoud20172d,qi2019deep,shi20193d,qi2018frustum}. Although those methods have achieved impressive results, the high-quality 3D ground truths are expensive and time-consuming to collect. Due to the capacity of alleviating the dependency on labeled data,  semi-supervised 3D object detection has drawn wide attention from researchers~\cite{zhao2020sess,wang20213dioumatch}. However, current 3D SSL detection works are all FixMatch-like and utilize a fixed threshold to select pseudo labels. None of them takes into account the variance of learning difficulty and class-imbalance situation, which leads to the sub-optimal results. In this work, a general class-level confidence based 3D SSL method is proposed to dynamically set the thresholds and re-sample the data according to the learning status. 

\begin{figure*}[t!]
\centering
\setlength{\belowcaptionskip}{-10pt}
\includegraphics[width = 0.9\textwidth]{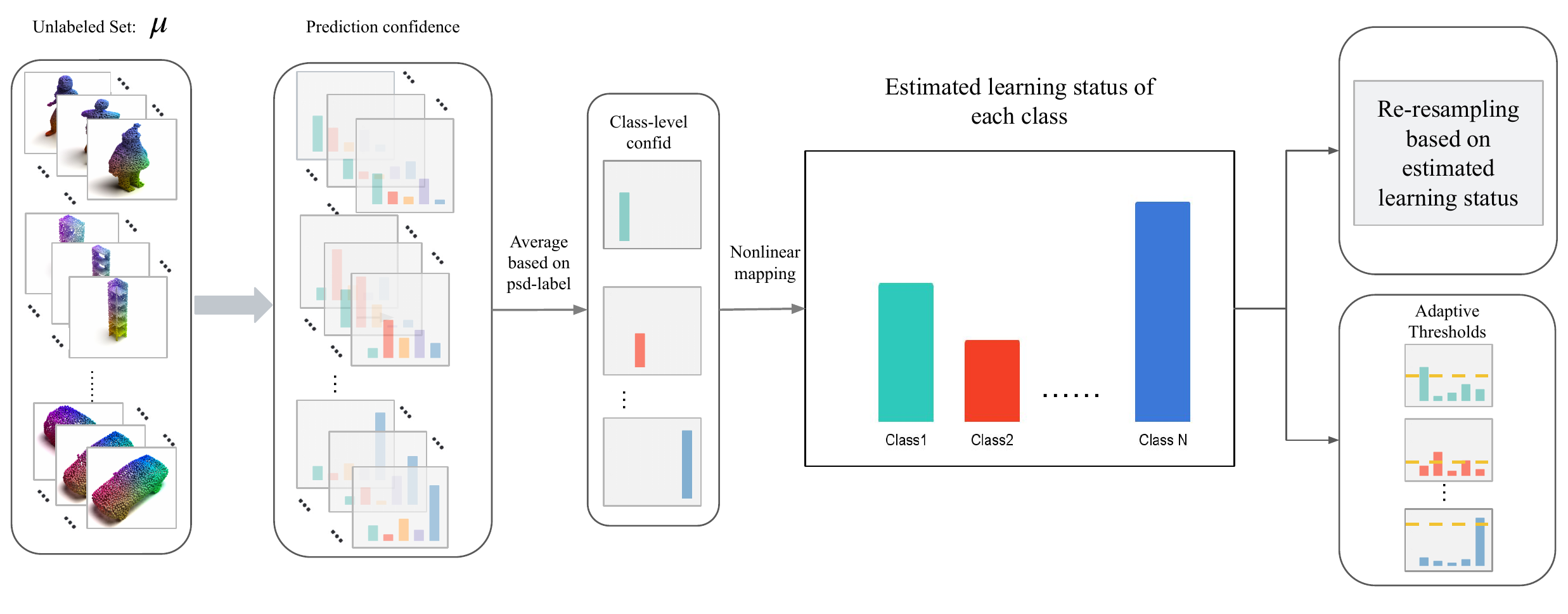}
\caption{\textbf{An overview of our proposed method}. Our model consists of three main parts: (1) Obtaining the learning status for each class through the class-level confidence from unlabeled data, (2) Leveraging learning status to dynamically adjusting the threshold of each class , and (3) Re-sampling the dataset based on the learning status.}
\label{fig:framework}
\end{figure*}

\section{Method}

The overview of our proposed method is shown in Fig~\ref{fig:framework}. It contains three main parts: (1) learning status estimation, which is obtained by the class-level confidence based on the model predictions on the unlabeled data, (2) dynamic thresholding based on the learning status of each lass, and (3) dynamically re-sampling for each class based on the learning status. The formulation for each component is introduced in the following sections.

\subsection{Problem Formulation}
The goal of the 3D semi-supervised learning is to jointly train the model based on limited labeled samples and a large number of unlabeled samples. Let $X_L = {(x_i,y_i)}_{i=1}^{N_L}$ be a limited labeled dataset with $N_l$ samples, where $x_i\in \mathbb{R}^{N\times3}$ is a 3D point cloud representation of an object or a scene, and $y_i$ is the corresponding label. Let $X_U = {(x_i)}_{i=1}^{N_U}$ be an unlabeled set with $N_U$ samples, which does not contain labels. Our model is trained on both $X_L$ and $X_U$ for semi-supervised learning with the proposed class-level confidence based dynamic threshold and re-sampling strategy.

\subsection{Baseline: FixMatch}
\label{sec:fixmatch}
Consistency regularization is a commonly used constraint in recent SSL algorithms ~\cite{berthelot2019mixmatch,berthelot2019remixmatch,tarvainen2017mean}. It forces prediction results from different augmentation of same instance being consistent:
\begin{equation}
\sum\limits_{b=1}^{\mu B}{||p_m(y|\alpha(u_b)) - p_m(y|\alpha(u_b))||}^2_2
\end{equation}, where $B$ represents the batch size of labeled data, $\mu$ is the labeled data and unlabeled data ratio, $\alpha$ is a stochastic data augmentation function, $u_b$ is the unlabeled data from $X_U$, and $p_m$ is the output probability of the model. Another popular method in SSL is pseudo-labeling, which obtain pseudo labels from unlabeled data prediction results. It defines a fixed threshold to cut off high-confidence unlabeled data to render pseudo labels. The cross-entropy loss is utilized to minimize the difference of the predictions and hard pseudo-labels:
\begin{equation}
\begin{aligned}
\frac{1}{\mu B} \sum\limits_{b=1}^{\mu B} \mathbb{1}(max(p_m(y|\alpha(u_b)) \geq \tau)\\ \cdot H(\hat{p}_m(y|\alpha(u_b),p_m(y|\alpha(u_b))
\end{aligned}
\end{equation}, where $\hat{p}_m(y|\alpha(u_b) = \arg\max(p_m(y|\alpha(u_b))$ and $\tau$ is the threshold. $H(p,q)$ represents the cross entropy loss between $p$ and $q$. Usually, a high threshold $\tau$ will be used to filter out low quality pseudo labels that have low prediction confidence.

Recently, FixMatch~\cite{sohn2020fixmatch} combines consistency regularization and pseudo-labeling together and achieved state-of-the-art performance on many tasks~\cite{wang20213dioumatch,zhao2020sess,chen2021multimodal}. FixMatch contains a supervised loss $\ell_s$ and a unsupervised loss $\ell_u$. The supervised loss is defined as:
\begin{equation}
\ell_s = \frac{1}{B} \sum\limits_{b=1}^{B}H(y_b,p_m(y|\alpha(x_b)) 
\end{equation}, where $y_b$ is the label of labeled instance $x_b$. For unsupervised loss $\ell_u$, FixMatch chooses the confident predictions (larger than the threshold) from weak augmented data as pseudo labels. Then, a cross-entropy loss is minimized based on the network prediction from the strong augmented views of the data and this pseudo label. The unsupervised loss is formulated as:
\begin{equation}
\begin{aligned}
\ell_u = \frac{1}{\mu B} \sum\limits_{b=1}^{\mu B} \mathbb{1}(max(p_m(y|\alpha(u_b)) \geq \tau)\\ \cdot H(\hat{p}_m(y|\alpha(u_b),p_m(y|\mathcal{A}{(u_b)})
\end{aligned}
\end{equation}
Where $\mathcal{A}$ is a strong augmentation function. Due to the simplicity and high performance of FixMatch, currently, most SSL methods are FixMatch-liked for many different tasks. However, in FixMatch and FixMatch-liked methods, the threshold $\tau$ is normally a high constant value. Although such a high threshold can improve the quality of pseudo labels, it will decrease the number of pseudo-labels that are actually used to optimize the network and leaves a large number of unlabeled data unused.

\subsection{Class-Level Confidence Based Dynamic Threshold}
\label{sec:threshold}



Inspired by FlexMatch ~\cite{zhang2021flexmatch}, we proposed a dynamic threshold based on the class-level confidence to leverage the learning status of each class. In existing SSL methods, instance-level prediction confidence is leveraged to evaluate the quality of the instance. However, class-level confidence remains to be un-utilized. Our analysis shows that for each class, their class-level confidence on the unlabeled data can be leveraged to represent the learning status. We can use the estimated learning status to dynamically adjust each class threshold and thus boost the effectiveness of SSL. For each class, we obtain its unlabeled set from prediction probability $argmax$: $\{ C_{c} | u_b \in C_{c}, \arg\max(p_m(y|\alpha(u_b)))=c, b = 1, 2, ..., \mu B \}$. Then, each unlabeled class set is averaged to obtain class-level confidence.
\begin{equation}
P_{c} = \frac{1}{|C_{c}|}\sum_{j=0}^{|C_{c}|} \max(p_m(y|\alpha(u_j)), u_j\in C_{c}
\end{equation}
Then, a non-linear function is utilized to map class-level confidence to the learning status. The entire process to obtain the dynamic formulated as:
\begin{equation}
\label{eq2}
\tau_{e}(c) =
\begin{cases}
 1 - \tau,  \hspace{0.1cm} & if \hspace{0.1cm}  M(P_{c}) < 1 - \tau  \\
 \tau,  \hspace{0.1cm} & elif \hspace{0.1cm}  M(P_{c}) > \tau \\
 M(P_{c}), \hspace{0.1cm} & else  \\
\end{cases}
\end{equation}
where $P_{c}$ is the class-level confidence for class $c$, $M(x) = \frac{x}{2-x}$ is a nonlinear mapping function, $\tau_{e}(c)$ is the dynamic threshold for class $c$ at epoch $e$. As there are no labels for unlabeled data, for each unlabeled instance $u_i$, the $argmax$ of the network prediction is utilized as its class: $c_{i}=\arg\max(p_m(y|\alpha(u_i)))$. The high-quality pseudo-labels are filtered by dynamic thresholds:
\begin{equation}
    \hat{y}_{c}^{i} = \mathbb{1}\left[ p_i\geq \tau_{e}(c_{i})\right]
\end{equation}
, where $p_i=\max(p_m(y|\alpha(u_i)))$.
The threshold for low learning statuses classes will be reduced to increase the pseudo label numbers and improve the learning status. As our proposed method adjusts thresholds method only based on class-level confidence, it can be applied to any kind of dataset and have higher generalization ability.The unsupervised loss of proposed dynamic thresholds method is formulated as:

\begin{equation}
\ell_{u,e}=\frac{1}{\mu B} \sum\limits_{b=1}^{\mu B} H(\hat{y}_{c}^{i}, p_m(y|\mathcal{A}{(u_b)})
\end{equation} 

\subsection{Class-Level Confidence based Data Re-sampling}
\label{sec:re-sampling}

While our proposed dynamic threshold can increase the pseudo-label numbers of low learning status classes, it still cannot completely balance each class's learning status due to data imbalance and the variance of learning difficulty. For example: in ModelNet40, the label numbers of airplane and bowl are 563 and 59 separately. Even if the dynamic threshold filters half pseudo-labels of airplane and utilizes all pseudo-labels of bowls, the airplane's selected unlabeled data numbers are at least four times larger than the bowl's selected unlabeled data numbers. Furthermore, the airplane has lower learning difficulty than the bowl due to its unique shape. Hence, even with dynamic threshold, classes like airplane still have high learning and tend to be overfitted due to their abundant data numbers, low learning difficulty, or both. This compromises the network performance. 

To alleviate this problem, we resort to the re-sampling strategy which has been proved to be effective in SSL class-imbalanced datasets. Most of current SSL imblanced method samples based on data numbers ~\cite{wei2021crest,he2021rethinking}. However, we find that some minority classes may have better performance than majority classes due to their low learning difficulty. Sampling based on data numbers makes the
model biased toward those low learning difficulty classes. 
Hence, we propose a class-level confidence based re-sampling strategy to directly increase the sampling probability of low learning status classes, which takes into account both learning difficulty and data imbalance. The sample probability for each class is formulated as:
\begin{equation}
\centering
\label{eq2}
\begin{cases}
1 - W(e)\cdot P_c \cdot p_i  & ,  if \hspace{0.1cm} P_c > \tau \hspace{0.2cm} , \\
2 - W(e)\cdot P_c \cdot p_i  & ,  if \hspace{0.1cm} P_c \leq \tau \hspace{0.2cm} , \\
\end{cases}
\end{equation}
, where the warm-function $W(e)=exp{(-5 \times (1 - e/E_{max})^2)}$ following previous works~\cite{yu2021playvirtual,ma2021adequacy} to avoid aggressively sampling. $p_i$ is the prediction confidence for instance $i$, $c$ is the prediction class for instance $i$, $P_c$ is the class-level confidence for class $c$, $e$ is the current epoch, and $E_{max}$ is the max epoch.

\subsection{Final Objective Function}
\label{sec:extend to general}
The core idea of our method is to dynamically adjust the pseudo-label threshold and re-sample the data based on learning status. Therefore, the proposed method can be easily applied to other pseudo-label based methods. Our entire model is jointly trained with two loss functions, including supervised loss $\ell_s$ on the labeled data and the $\ell_{u,e}$ on the unlabeled data as:
\begin{equation}
    \ell = \lambda_s\ell_s + \lambda_u\ell_{u,e}
\end{equation}
,where $\lambda_s$ and $\lambda_u$ are weights for labeled and unlabeled losses.

\section{Experimental Results}

\begin{table*}[http]
    \centering

\adjustbox{width=400pt,height=620pt,keepaspectratio=true}
{

	\begin{tabular}{K{15mm}|K{15mm}|K{15mm}|K{15mm}|K{15mm}|K{15mm}|K{15mm}|K{15mm}|K{15mm}|K{15mm}}
	    \hline
        \multicolumn{2}{c|}{\multirow{3}{*}{Dataset}}&\multicolumn{2}{c|}{\multirow{3}{*}{Method}} &\multicolumn{2}{c|}{2\% } &\multicolumn{2}{c|}{5\% } &\multicolumn{2}{c}{10\% } 
        
        \\\cline{5-10}
        \multicolumn{2}{c|}{} & \multicolumn{2}{c|}{} & Overall & Mean& Overall& Mean& Overall& Mean
        \\
        \multicolumn{2}{c|}{} & \multicolumn{2}{c|}{} & Acc & Acc& Acc& Acc& Acc& Acc\\

        \hline
        \multicolumn{2}{c|}{\multirow{7}{*}{ ModelNet40}} & \multicolumn{2}{c|}{Point Transformer\cite{guo2021pct}} & 71.1 & 61.0 & 77.1& 69.2& 84.6& 77.2  
        \\\cline{3-10}
        \multicolumn{2}{c|}{} & \multicolumn{2}{c|}{PL\cite{lee2013pseudo}} & 69.7  & 59.6 & 78.3& 69.0& 85.1& 77.7
        \\\cline{3-10}
        \multicolumn{2}{c|}{} &
        \multicolumn{2}{c|}{Flex-PL\cite{zhang2021flexmatch}} & 66.7 & 54.9 & 74.2& 62.3& 83.2& 70.3
        \\\cline{3-10}
        \multicolumn{2}{c|}{} &
        \multicolumn{2}{c|}{Confid-PL(Ours)} & \textbf{74.4} & \textbf{61.9} & \textbf{80.6}& \textbf{73.5}& \textbf{86.5}& \textbf{80.4}
        \\\cline{3-10}
        \multicolumn{2}{c|}{} & \multicolumn{2}{c|}{FixMatch\cite{sohn2020fixmatch}(NeurIPS 2020)} & 70.8 & 62.7 & 78.9& 71.1& 85.5& 79.4
        \\\cline{3-10}
        \multicolumn{2}{c|}{} & \multicolumn{2}{c|}{Dash\cite{xu2021dash}(ICML 2021)} & 71.5 & 63.0 & 79.7& 71.8& 85.9& 80.1
        \\\cline{3-10}
        \multicolumn{2}{c|}{} & \multicolumn{2}{c|}{FlexMatch\cite{zhang2021flexmatch}(NeurIPS 2021)} & 70.1 & 61.2 & 80.5& 70.4& 86.2& 78.7
        \\\cline{3-10}
        \multicolumn{2}{c|}{} & \multicolumn{2}{c|}{Confid-Match(Ours)} & \textbf{73.8} & \textbf{64.1} & \textbf{82.1}& \textbf{74.3}& \textbf{87.8}& \textbf{82.5}
        \\
        \hline
	\end{tabular}
}
\caption{Comparison with state-of-the-art methods. The results on the ModelNet40 dataest for 3D semi-supervised object classification task.}
\label{tab:Classification-modelnet}
\end{table*}

\begin{table*}[http]
    \centering

\adjustbox{width=400pt,height=620pt,keepaspectratio=true}
{

	\begin{tabular}{K{15mm}|K{15mm}|K{15mm}|K{15mm}|K{15mm}|K{15mm}|K{15mm}|K{15mm}|K{15mm}|K{15mm}}
	    \hline
        \multicolumn{2}{c|}{\multirow{3}{*}{Dataset}}&\multicolumn{2}{c|}{\multirow{3}{*}{Method}} &\multicolumn{2}{c|}{1\% } &\multicolumn{2}{c|}{2\% } &\multicolumn{2}{c}{5\% } 
        
        \\\cline{5-10}
        \multicolumn{2}{c|}{} & \multicolumn{2}{c|}{} & Overall & Mean& Overall& Mean& Overall& Mean
        \\
        \multicolumn{2}{c|}{} & \multicolumn{2}{c|}{} & Acc & Acc& Acc& Acc& Acc& Acc\\
        
        \hline
        \multicolumn{2}{c|}{\multirow{7}{*}{ScanObjectNN}} & \multicolumn{2}{c|}{Point Transformer\cite{guo2021pct}} & 32.1 & 26.1 & 44.7& 36.5& 56.6& 50.0  
        \\\cline{3-10}
        \multicolumn{2}{c|}{} & \multicolumn{2}{c|}{PL\cite{lee2013pseudo}} & 31.2 & 25.8 & 47.5& 38.6& 58.1& 51.5
        \\\cline{3-10}
        
        \multicolumn{2}{c|}{} &
        \multicolumn{2}{c|}{Flex-PL\cite{zhang2021flexmatch}} & 29.2 & 24.2 & 47.2& 37.6& 60.1& 51.9
        \\\cline{3-10}
        \multicolumn{2}{c|}{} &
        \multicolumn{2}{c|}{Confid-PL(Ours)} & \textbf{32.6} & \textbf{27.1} & \textbf{48.8}& \textbf{41.5}& \textbf{63.1}& \textbf{55.2}
        \\\cline{3-10}
        
        \multicolumn{2}{c|}{} & \multicolumn{2}{c|}{FixMatch\cite{sohn2020fixmatch}(NeurIPS 2020)} & 33.5 & 27.6 & 47.4& 39.9& 59.4& 52.4
        \\\cline{3-10}
        \multicolumn{2}{c|}{} & \multicolumn{2}{c|}{Dash\cite{xu2021dash}(ICML 2021)} & 35.1 & 29.3 & 50.3& 44.1& 62.8& 60.3
        \\\cline{3-10}
        \multicolumn{2}{c|}{} & \multicolumn{2}{c|}{FlexMatch\cite{zhang2021flexmatch}(NeurIPS 2021)} & 34.2 & 26.2 & 48.5& 39.7& 63.4& 57.2
        \\\cline{3-10}
        \multicolumn{2}{c|}{} & \multicolumn{2}{c|}{Confid-Match(Ours)} & \textbf{38.2} & \textbf{32.7} & \textbf{57.0}& \textbf{48.6}& \textbf{69.4}& \textbf{65.5}
        \\
        \hline

	\end{tabular}
}\caption{Comparison with state-of-the-art methods. The results on the ScanObjectNN dataest for 3D semi-supervised object classification task.}
\label{tab:Classification-objectnn}
\end{table*}

\subsection{Datasets}
\textbf{Classification Datasets:} Following the state-of-the-art SSL 3D object classification methods, we evaluate our method on two benchmarks including ScanObjectNN \cite{uy-scanobjectnn-iccv19} and ModelNet40~\cite{wu20153d}. The ModleNet40 is a widely used benchmark and it contains $12,311$ meshed CAD models from $40$ categories, and there are $9,843$ models in training and $2,468$ in testing. The ScanObjectNN is a more realistic point cloud object dataset. It contains $15,000$ objects belongs to 15 classes that sampled from $2,902$ unique object instances from real world.

\textbf{Detection Datasets:} Following the previous state-of-the-art SSL 3D object detection methods~\cite{wang20213dioumatch}, we evaluate our method on two widely used detection benchmark including SUN RGB-D~\cite{song2015sun} and ScanNet~\cite{dai2017scannet}. The ScanNet~\cite{dai2017scannet} is an indoor scene dataset consisting of $1513$ reconstructed meshes, among which $1201$ are training samples, and $312$ are validation samples. SUN RGB-D~\cite{song2015sun} contains more than $10,000$ indoor scenes while $5285$ for training and $5050$ for validation.

\subsection{Implementation Details}
\textbf{Semi-Supervised 3D Object Classification:} On the ModelNet40 and ScanObjectNN datasets, we use SGD optimizer with a learning rate of $0.01$, and the learned rate is scheduled with CosineAnnealingLR decay with a minimum learning rate of $0.0001$. All the models are optimized with a total epoch of $500$. The weak augmentation contains rotation and random scale, and the strong augmentation leverages random scale, translation, jittering, and rotation. The batch size is set to $240$, while 48 of them are labeled data and the rest are unlabeled. Weights of supervised loss and unsupervised loss are both $1$. The threshold is set to be $\tau = 0.8$. The re-sampling strategy updates the dataloader every $50$ epoch. The PointTransformer~\cite{guo2021pct} is utilized as the backbone for the SSL classification task.

\textbf{Semi-Supervised 3D Object Detection:} 
We apply our method on state-of-the-art work 3DIoUMatch and follow the same setting. Unlike 3DIoUMatch uses VoteNet pre-trained model, we utilize the proposed re-sampling strategy to re-sample the labeled data in the pre-training process. Then the pre-trained weights are utilized to initialize the student and teacher networks. For those multiple objects sceneries, the object that has minimal confidence is leveraged in the re-sampling process. Like classification task, the re-sampling strategy updates the dataloader every $50$ epoch. For a fair comparison, the pre-processing data methods and labels are the same as previous works ~\cite{qi2019deep,wang20213dioumatch} and mean average precision (mAP) under IoU thresholds of 0.25 and 0.5 are utilized as evaluation metrics. The threshold is also set to be $\tau = 0.8$.

\begin{table*}[t!]
\setlength{\belowcaptionskip}{-2pt}
    \centering

\adjustbox{width=480pt,height=800pt,keepaspectratio=true}
{

	\begin{tabular}{K{15mm}|K{15mm}|K{15mm}|K{15mm}|K{15mm}|K{15mm}|K{15mm}|K{15mm}|K{15mm}|K{15mm}}
	    \hline
        \multicolumn{2}{c|}{\multirow{3}{*}{Dataset}}&\multicolumn{2}{c|}{\multirow{3}{*}{Method}} &\multicolumn{2}{c|}{1\% } &\multicolumn{2}{c|}{2\% } &\multicolumn{2}{c}{5\% } 
        
        \\\cline{5-10}
        \multicolumn{2}{c|}{} & \multicolumn{2}{c|}{} & mAP & mAP& mAP& mAP& mAP& mAP
        \\
        \multicolumn{2}{c|}{} & \multicolumn{2}{c|}{} & $@$ 0.25 & $@$0.50& $@$0.25& $@$0.50& $@$0.25& $@$0.50\\
        \hline
        \multicolumn{2}{c|}{\multirow{4}{*}{SUN  RGB-D}} & \multicolumn{2}{c|}{VoteNet~\cite{qi2019deep}(ICCV 2019)} & 16.7$\pm$1.2 & 3.9$\pm$0.9 & 21.8$\pm$1.6& 5.1$\pm$0.
        8& 33.9$\pm$1.9& 13.1$\pm$1.7 
        \\\cline{3-10}
        \multicolumn{2}{c|}{} & \multicolumn{2}{c|}{SESS~\cite{zhao2020sess}(CVPR 2020)} & 19.9$\pm$1.6 & 6.3$\pm$1.2 &23.3$\pm$1.1 & 7.9$\pm$0.8& 36.1$\pm$1.1& 16.9$\pm$0.9
        \\\cline{3-10}
        \multicolumn{2}{c|}{} & \multicolumn{2}{c|}{3DIoUMatch ~\cite{wang20213dioumatch} (CVPRR 2021)} 
        & 25.6$\pm$0.6 
        & 9.4$\pm$0.7 
        & 26.8$\pm$0.7
        & 10.6$\pm$0.5
        & 39.7$\pm$0.9
        & 20.6$\pm$0.7
        \\\cline{3-10}
        \multicolumn{2}{c|}{} & \multicolumn{2}{c|}{Confid-3DIoUMatch(Ours)} & \textbf{27.8$\pm$0.8} & \textbf{11.3$\pm$0.6} & \textbf{32.7$\pm$0.3}& \textbf{13.5$\pm$0.4}& \textbf{43.1$\pm$0.6}& \textbf{24.2$\pm$0.5}
        \\
        \hline

	\end{tabular}
}

 \caption{Comparative studies with state-of-the-art methods on the SUN RGB-D dataest for 3D SSL object detection.}
\label{tab:Detection-sun}
\end{table*}

\begin{table*}[]
\setlength{\belowcaptionskip}{-10pt}
    \centering

\adjustbox{width=480pt,height=800pt,keepaspectratio=true}
{

	\begin{tabular}{K{15mm}|K{15mm}|K{15mm}|K{15mm}|K{15mm}|K{15mm}|K{15mm}|K{15mm}|K{15mm}|K{15mm}}
	    \hline
        \multicolumn{2}{c|}{\multirow{3}{*}{Dataset}}&\multicolumn{2}{c|}{\multirow{3}{*}{Method}} &\multicolumn{2}{c|}{1\% } &\multicolumn{2}{c|}{2\% } &\multicolumn{2}{c}{5\% } 
        
        \\\cline{5-10}
        \multicolumn{2}{c|}{} & \multicolumn{2}{c|}{} & mAP & mAP& mAP& mAP& mAP& mAP
        \\
        \multicolumn{2}{c|}{} & \multicolumn{2}{c|}{} & $@$ 0.25 & $@$0.50& $@$0.25& $@$0.50& $@$0.25& $@$0.50\\
        \hline
        \multicolumn{2}{c|}{\multirow{4}{*}{ScanNet}} & \multicolumn{2}{c|}{VoteNet~\cite{qi2019deep}(ICCV 2019)} & $8.9\pm1.1$ & $1.5\pm0.5$ & $16.9\pm1.3$& $4.7\pm0.8$& $31.2\pm1.1$& $14.7\pm0.7$  
        \\\cline{3-10}
        \multicolumn{2}{c|}{} & \multicolumn{2}{c|}{SESS~\cite{zhao2020sess}(CVPR 2020)} &$11.3\pm1.6$&$2.7\pm0.6$& $21.1\pm1.5$&$8.4\pm1.1$&$35.5\pm2.0$&$17.2\pm0.9$
        \\\cline{3-10}
        \multicolumn{2}{c|}{} & \multicolumn{2}{c|}{3DIoUMatch ~\cite{wang20213dioumatch} (CVPRR 2021)} & $14.6\pm1.4$&$3.9\pm0.5$ & $24.5\pm1.9$& $11.2\pm1.4$&$40.4\pm0.8$&$21.0\pm0.6$
        \\\cline{3-10}
        \multicolumn{2}{c|}{} & \multicolumn{2}{c|}{Confid-3DIoUMatch(Ours)} & \textbf{19.0$\pm$0.4} & \textbf{6.4$\pm$0.4} & \textbf{29.5$\pm$1.5}& \textbf{15.2$\pm$0.6}& \textbf{43.6$\pm$0.5}& \textbf{24.3$\pm$0.4}
        \\
        \hline

	\end{tabular}
}
 \caption{Comparative studies with state-of-the-art methods on the ScanNet dataest for 3D SSL object detection.}
\label{tab:Detection-scan}
\end{table*}



\subsection{Performance on Semi-Supervised 3D Object Classification}
To demonstrate the capability and potential of our proposed method, we compare the performance of our method with other semi-supervised learning methods including Pseudo-Labeling (PL)~\cite{lee2013pseudo}, FixMatch~\cite{sohn2020fixmatch}, Dash~\cite{xu2021dash}, and FlexMatch~\cite{zhang2021flexmatch}) under the same setting in ModelNet40~\cite{wu20153d} and ScanObjectNN~\cite{uy-scanobjectnn-iccv19}. The FlexMatch~\cite{zhang2021flexmatch} is the most recent state-of-the-art method that was proposed to overcome the fixed-threshold drawback of FixMatch by proposing curriculum pseudo labeling to dynamically adjust the threshold. For a fair comparison, all the methods use the same backbone, data augmentation, and hyper-parameters.

To extensively compare with the state-of-the-art methods, we apply our method in FixMatch and Pseudo-labeling and report the results for all the methods under different percentages of labeled data. Two evaluation metrics are used to indicate the performance, including overall accuracy and class mean accuracy, which is the average of the accuracy of all the classes. The Table~\ref{tab:Classification-modelnet} and Table~\ref{tab:Classification-objectnn} indicate that in the 3D SSL clasification task, the current state-of-the-art work FlexMatch only has limited improvement or even \textbf{decreases} the performance when applied in FixMatch and Pseudo-Labeling with limited labeled data.  This is because FlexMatch is only designed for class balanced dataset but 3D datasets are all inherently data-imbalanced. As shown in Table~\ref{tab:Classification-modelnet}, our model outperforms all the state-of-the-art methods with two evaluation metrics on the ModelNet40 dataset. As shown in Table~\ref{tab:Classification-objectnn}, on more realistic and challenging dataset ScanObjectNN~\cite{uy-scanobjectnn-iccv19}, our model also significantly outperforms all the other methods by a large margin. For both ModelNet40~\cite{wu20153d} and ScanObjectNN~\cite{uy-scanobjectnn-iccv19} dataset, the improvement of our model on the mean class accuracy is more significant, which demonstrates the effectiveness of our model on balancing the learning status of each class.

\subsection{Performance on Semi-Supervised 3D Object Detection}

To demonstrate the generalization ability of our method, we further evaluated our proposed method on the semi-supervised 3D object detection benchmark and compared it with the state-of-the-art methods. We extend the state-of-the-art method 3DIoUMatch~\cite{wang20213dioumatch} and apply our class-based thresholding and re-sampling during training. We report the performance comparison with the state-of-the-art methods including VoteNet~\cite{qi2019deep}, SESS~\cite{zhao2020sess} and 3DIoUMatch~\cite{wang20213dioumatch} on two benchmark including SUN RGB-D and ScanNet datasets. Following the convention, the mean average precision (mAP) under two different thresholds, including $0.25$ and $0.5$ are reported.

As shown in Table~\ref{tab:Detection-sun} and~\ref{tab:Detection-scan}, our model significantly outperforms all the other state-of-the-art methods on both SUN RGB-D and ScanNet datasets with different settings. The most significant improvement is under 2\% labeled setting in which our method outperforms 3DIoUMatch by \textbf{5.9} and \textbf{5.0} on mAP@0.25 on ScanNet and SUN RGB-D, respectively. The results on those two benchmarks demonstrate that our proposed confidence-based dynamic threshold and learning status balance re-sampling strategy can be easily integrated into other semi-supervised methods to significantly boost the performance.



\begin{table*}[t!]
\setlength{\belowcaptionskip}{-10pt}
\centering
\adjustbox{width=300pt,height=260pt,keepaspectratio=true}
{
	\begin{tabular}{c|c|c|c|c|c}

	    \hline
	    \multirowcell{2}{Confidence \\Re-sample} & \multirowcell{2}{ Dynamic \\Threshold} & 
	    \multicolumn{2}{c|}{ModelNet40 10\%} &
	    \multicolumn{2}{c}{ObjectNN 5\%} \\ \cline{3-6}
	   && Overall Acc & Mean Acc & Overall Acc & Mean Acc
        \\
        \hline
         &  & 85.5& 79.4  & 59.4 &52.4   \\
        \hline
        \checkmark&  & 86.7  &81.1 & 64.1 & 60.5   \\
        \hline
        &  \checkmark & 86.9  & 81.7& 66.9 &  62.5  \\
        \hline

        \checkmark& \checkmark & \textbf{87.8} & \textbf{82.5} & \textbf{69.4} &  \textbf{65.5}    \\
        \hline
	\end{tabular}
}
 \caption{Ablation  study  for components effect on the 3D SSL object classification task in ModelNet40 and ScanObjctNN dataset.}
\label{tab:Classification_abl}
\end{table*}
\begin{table*}[t!]
\setlength{\belowcaptionskip}{-10pt}

\centering
\adjustbox{width=360pt,height=500pt,keepaspectratio=true}
{
	\begin{tabular}{c|c|c|c|c|c|c}

	    \hline
	    \multirowcell{2}{Re-sample\\Pre-train} &
	    \multirowcell{2}{Confidence \\Re-sample} & \multirowcell{2}{ Dynamic \\Threshold} & 
	    \multicolumn{2}{c|}{ScanNet 5\%} &
	    \multicolumn{2}{c}{SUN-RGBD 2\%} \\ \cline{4-7}
	    & & & mAP @0.25 & mAP @0.5 & mAP @0.25 & mAP @0.5
        \\

        \hline
        & &  & 40.4$\pm$0.8&21.0$\pm$0.6& 26.8$\pm$1.1 & 10.6$\pm$0.5  \\
        \hline
        \checkmark& &  &41.8$\pm$0.6 & 22.7$\pm$0.5  & 31.0$\pm$0.7 & 11.7$\pm$0.6  \\
        \hline
        \checkmark&& \checkmark &  42.1$\pm$0.8 & 23.2$\pm$0.5 & 31.7$\pm$0.8 &  12.4$\pm$0.5  \\
        \hline
        \checkmark&\checkmark&  &  42.4$\pm$0.4 & 23.0$\pm$0.6 & 31.9$\pm$0.9 &  11.9$\pm$0.7  \\
        \hline
        \checkmark& \checkmark& \checkmark & \textbf{43.6$\pm$0.5} & \textbf{24.3$\pm$0.4} & \textbf{32.7$\pm$0.3}&\textbf{13.5$\pm$0.4}      \\
        \hline

	\end{tabular}
}
 \caption{Ablation  study  for  components effect on the 3D SSL object detection task in ScanNet and Sun RGB-D dataset.}
\label{tab:detection_abl}
\end{table*}

\subsection{Ablation Study for Dynamic Threshold and Re-sampling}

Our proposed method contains two major components: confidence-based dynamic threshold and dynamic re-sampling strategy. To analyze the effect of each component, we conduct ablation studies about the combinations of different components on both SSL 3D object classification and detection tasks. The FixMatch~\cite{sohn2020fixmatch} is used as baseline for classification task while 3DIoUMatch~\cite{wang20213dioumatch} is used as baseline for detection task. The Table~\ref{tab:Classification_abl} contains the results for classification while Table~\ref{tab:detection_abl} is for detection task.

\textbf{Dynamic Threshold:} For both the classification and detection tasks, the baseline uses the fixed threshold to select unlabeled data with high-quality predictions. After applying our class-level dynamic threshold strategy to the two different tasks, the performance is improved for both tasks under different settings. The improvement for some settings is huge, i.e. the mean accuracy for ScanObjectNN dataset is improved by $10.1$\% under the $5$\% labeled set. These results confirm the effectiveness of our class-level dynamic threshold and show that our method can be easily integrated into other semi-supervised learning methods.

\textbf{Dynamic Re-sampling:} For both the classification and detection, the baselines do not use any re-sampling strategy, therefore, the sampling probability for each data sample are same. For the SSL 3D classification task, all the performances are improved after applying our proposed dynamic re-sampling strategy, and the improvement on the realistic dataset ObjectNN is the most significant. For the SSL 3D detection task, performing the re-sampling strategy during the pre-train stage to re-sample labeled data only can significantly improve the performance, while the performance can be further improved by also applying the re-sampling during the semi-supervised training stage. The results on both SSL 3D classification and detection benchmarks show the effectiveness of the proposed re-sampling strategy.

\subsection{Ablation Study to Other Design Choice}

To better understand our method, we conduct ablation studies to evaluate the impact of upper limit of thresholds and mapping functions. To comprehensively evaluate our method, we provide the ablation results on both SSL 3D classification task with $10$ percent labeled data in ModelNet40 dataset and SSL 3D object detection task with $5$ percent labeled data in ScanNet dataset.

\textbf{Upper Limit Threshold:} To investigate the impact of the upper limit threshold on our proposed method,  we conduct ablation experiments on both classification and detection tasks. The results for SSL 3D classification and detection tasks are shown in Table.~\ref{tab:classification-threshold-abs} and Table.~\ref{tab:detection-threshold-abs} respectively. The $0.8$ threshold achieves the best performance for both classification and detection tasks.

\textbf{Learning Status Mapping Function:}
To better understand the effect of learning status mapping function, we verified the results of three different mapping functions including: (1) exponential: $M(x_c) = exp{(-5 \times (1 - P_c)^2)}$ (2) linear: $M(x_c) = P_c$, and (3) concave: $M(_c)=P_c/(2-P_c)$. Where $P_c$ is the class-level confidence for class $c$. The results for SSL 3D classification and detection tasks are shown in  Table.~\ref{tab:classification-mapping-abs} and Table.~\ref{tab:detection-mapping-abs} respectively. We can see that the concave function leads to best performance for both classification and detection tasks. Besides, the performance of three mapping functions are similar for both tasks, which indicates the robustness of our method.

\begin{table}[t!]
    \begin{subtable}[h]{0.2\textwidth}
         \centering
        \begin{tabular}{c|cc}
            \multirow{2}{*}{$\tau$} & Overall & Mean \\  {}& Acc & Acc\\\hline
            $0.75$ &$87.0$ & $81.5$\\
            $0.8$ & $\textbf{87.8}$& $82.5$\\
            $0.85$ & $87.5$& $\textbf{83.1}$\\
        \end{tabular}
        \captionsetup{width=0.9\linewidth}
        \caption{Upper limit threshold for classification.}
        \label{tab:classification-threshold-abs}
    \end{subtable}
    \hspace{3mm}
    \begin{subtable}[h]{0.2\textwidth}
         \centering
        \begin{tabular}{c|cc}
            \multirowcell{2}{Mapping\\function} & Overall & Mean \\  {}& Acc & Acc\\\hline
            Concave & $\textbf{87.8}$& $\textbf{82.5}$\\
            Linear & $87.1$& $81.4$\\
            Exp & $87.4$& $82.1$\\
        \end{tabular}
        \captionsetup{width=0.9\textwidth}
        \caption{Mapping function for classification.}
        \label{tab:classification-mapping-abs}
    \end{subtable}
    
    \begin{subtable}[h]{0.2\textwidth}
         \centering
        \begin{tabular}{c|cc}
              \multirow{2}{*}{$\tau$} & mAP & mAP \\  {}& @0.25 & @0.5\\\hline
            $0.75$ &$42.7$ &$23.6$ \\
            $0.8$ &$\textbf{43.6}$ & $\textbf{24.3}$\\
            $0.85$ & $43.2$& $24.1$\\
        \end{tabular}
        \captionsetup{width=.85\linewidth}
        \caption{Upper  limit threshold for detection.}
        \label{tab:detection-threshold-abs}
    \end{subtable}
    \hspace{3mm}
    \begin{subtable}[h]{0.2\textwidth}
         \centering
        \begin{tabular}{c|cc}
            \multirowcell{2}{Mapping\\function}  & mAP & mAP \\  {}& @0.25 & @0.5\\\hline
            Concave &$43.6$ & $\textbf{24.3}$\\
            Linear & $\textbf{43.8}$& $23.5$\\
            Exp & $43.0$& $24.0$\\
        \end{tabular}
        \captionsetup{width=.85\linewidth}
        \caption{Mapping function for detection.}
        \label{tab:detection-mapping-abs}
    \end{subtable}



    \caption{Ablation study for upper limit threshold and mapping function in 3D SSL classification and detection task. (a) and (b) are classification task. (d) and (e) are results from detection task. All the results confirm that our method is very robust to those design choices.}
    \label{tab:array}
\end{table}

\section{Conclusion}

In this work, we propose a novel class-level confidence based dynamic threshold method and re-sampling strategy. The proposed method not only improves the performance, but makes the prediction results balanced among classes. Our proposed method remarkably outperforms the state-of-the-art SSL classification and detection methods. These results demonstrate the effectiveness and generality of the proposed method in 3D SSL tasks. In the future, we will extend our method to 2D SSL tasks.

{\small
\bibliographystyle{ieee_fullname}
\bibliography{egbib}
}

\end{document}


\title{Supplementary Materials for Class-Level Confidence Based 3D Semi-Supervised Learning}

\author[1]{Zhimin Chen}
\author[2]{Longlong Jing}
\author[2]{Liang Yang}
\author[3]{Yingwei Li}
\author[1]{Bing Li}
\affil[1]{Clemson University} 
\affil[2]{The City University of New York}
\affil[3]{Johns Hopkins University}
\affil[ ]{\textit {\tt\small \{zhiminc,bli4\}@clemson.edu, ljing@gradcenter.cuny.edu, lyang1@ccny.cuny.edu, yingwei.li@jhu.edu}}
\thispagestyle{empty}
\maketitle

This document contains the supplementary materials for "Class-Level Confidence Based 3D Semi-Supervised Learning".

\section{Analysis on SSL Object Detection}
\begin{figure}[t!]
\centering
\includegraphics[width = 0.45\textwidth]{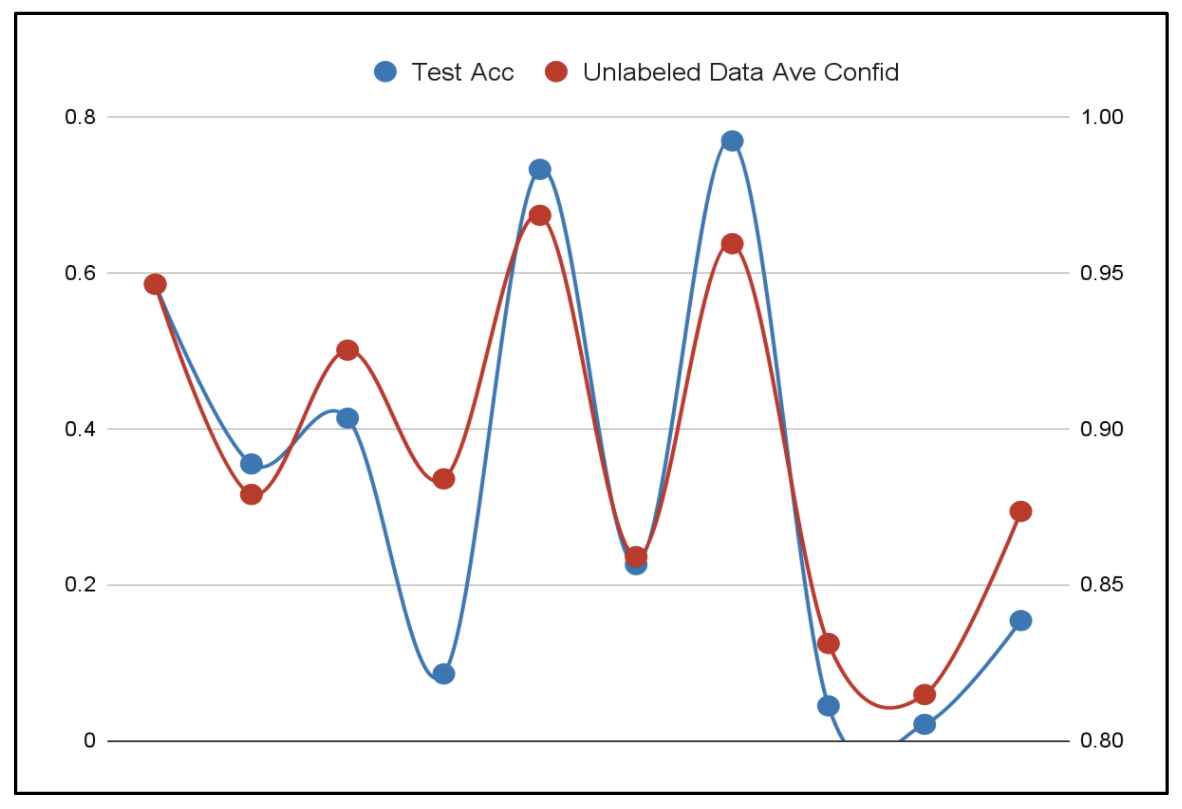}
\caption{The results analysis of 3DIoUMatch trained in SUN-RGBD dataset with 5\% labeled data. It indicates that the class-Level confidence has high correlation with test accuracy of each class in 3D SSL detection task.} 
\label{fig:det-correlation}
\end{figure}

In the main paper, we demonstrate the high correlation between class-level confidence of unlabeled data and test accuracy in 3D SSL classification task (line 100 to 104). Therefore, we hypothesize that class-level confidence of unlabeled data can be utilized to estimate learning status. To validate the generality of this correlation, we utilize 3DIoUMatch to conduct 3D SSL object detection experiment in SUN-RGBD dataset with $5$ percent dataset as shown in the Fig~\ref{fig:det-correlation}. The results demonstrate that the class-level confidence also has high correlation with test accuracy in detection tasks, which supports the generality of our hypothesis.

\section{Visualization of SSL Object Detection Results}
To take a deeper look at the prediction results of our model, we compared the qualitative results of the supervised baseline, the 3DIoUMatch~\cite{wang20213dioumatch}, and our results. The supervised baseline VoteNet~\cite{qi2019deep} produces many false positive predictions due to the very limited labeled data during training. The results of 3DIoUMatch are much better than the supervised baseline but still have many false positive boxes. Compared to the baseline and 3DIoUMatch, the quality of our method is much higher and with fewer false-positive boxes demonstrating the effectiveness of our method.

\subsection{Comparison with Class Imbalanced SSL Method}
Most of current SSL imblanced method resamples based on data numbers. However, we find that some minority classes may have better performance than majority classes due to their low learning difficulty. Sampling based on data numbers makes the model biased
toward those low learning difficulty classes. Unlike previous method,our re-sampling strategy directly increases the sampling probability of low learning status classes to balance the learning status. To further show the effectiveness of our method, we compare our method with recent state-of-the-art method BiS ~\cite{he2021rethinking} that relies on class cardinality to sample. Table.~\ref{tab:imbalance_compa_cls} and Table.~\ref{tab:imbalance_compa_det} indicate that our method still achieves better performance than BiS in both 3D detection and classification tasks when only the sampling part is utilized.

\begin{table}[http]
\centering
\adjustbox{width=240pt,height=360pt,keepaspectratio=true}
{
	\begin{tabular}{c|c|c|c|c}

	    \hline
	     \multirowcell{2}{ } & 
	    \multicolumn{2}{c|}{ModelNet40 5\%} &
	    \multicolumn{2}{c}{ModelNet40 10\%} \\ \cline{2-5}
	   & Overall Acc & Mean Acc & Overall Acc & Mean Acc
        \\
        \hline
        Baseline & 78.9 & 71.1 & 85.5& 79.4     \\
        \hline
        BiS~\cite{he2021rethinking} + Baseline & 79.7 & 72.3 & 86.1  &80.3 \\
        \hline
        Confid-Sample + Baseline & 81.0 & 72.8 & 86.7  &81.1   \\
        \hline
        Ours + Baseline & \textbf{82.1} & \textbf{74.3} & \textbf{87.8} & \textbf{82.5}     \\
        \hline
	\end{tabular}
}
 \caption{\small Comparative studies with state-of-the-art class imbalanced SSL method for 3D object classification.}
\label{tab:imbalance_compa_cls}
\end{table}

\begin{table}[http]
\centering
\adjustbox{width=240pt,height=360pt,keepaspectratio=true}
{
	\begin{tabular}{c|c|c|c|c}

	    \hline
	     \multirowcell{2}{ } & 
	    \multicolumn{2}{c|}{SUN RGB-D 2\%} &
	    \multicolumn{2}{c}{SUN RGB-D 5\%} \\ \cline{2-5}
	   & mAP $@$0.25 & mAP $@$0.5 & mAP $@$0.25 & mAP $@$0.5
        \\
        \hline
        Baseline &26.8 ± 1.1 &10.6 ± 0.5 & 39.7 ± 0.9 & 20.6 ± 0.7   \\
        \hline
        BiS~\cite{he2021rethinking} + Baseline & 29.2 ± 0.7 & 11.5 ± 0.4 & 41.5 ± 1.1 & 22.4 ± 0.8   \\
        \hline
        Confid-Sample + Baseline  &  31.9 ± 0.9 & 11.9 ± 0.7 & 42.4 ± 0.9 & 23.1 ± 0.6   \\
        \hline
        Ours + Baseline & \textbf{32.7 ± 0.3}  &  \textbf{13.5 ± 0.4}  & \textbf{43.1 ± 0.6}  &  \textbf{24.2 ± 0.5}     \\
        \hline
	\end{tabular}
}
 \caption{\small Comparative studies with state-of-the-art class imbalanced SSL method for 3D object detection.}
\label{tab:imbalance_compa_det}
\end{table}
\vspace{5mm}

\begin{figure*}[t!]
\centering
\setlength{\belowcaptionskip}{-10pt}
\includegraphics[width = 1.0\textwidth]{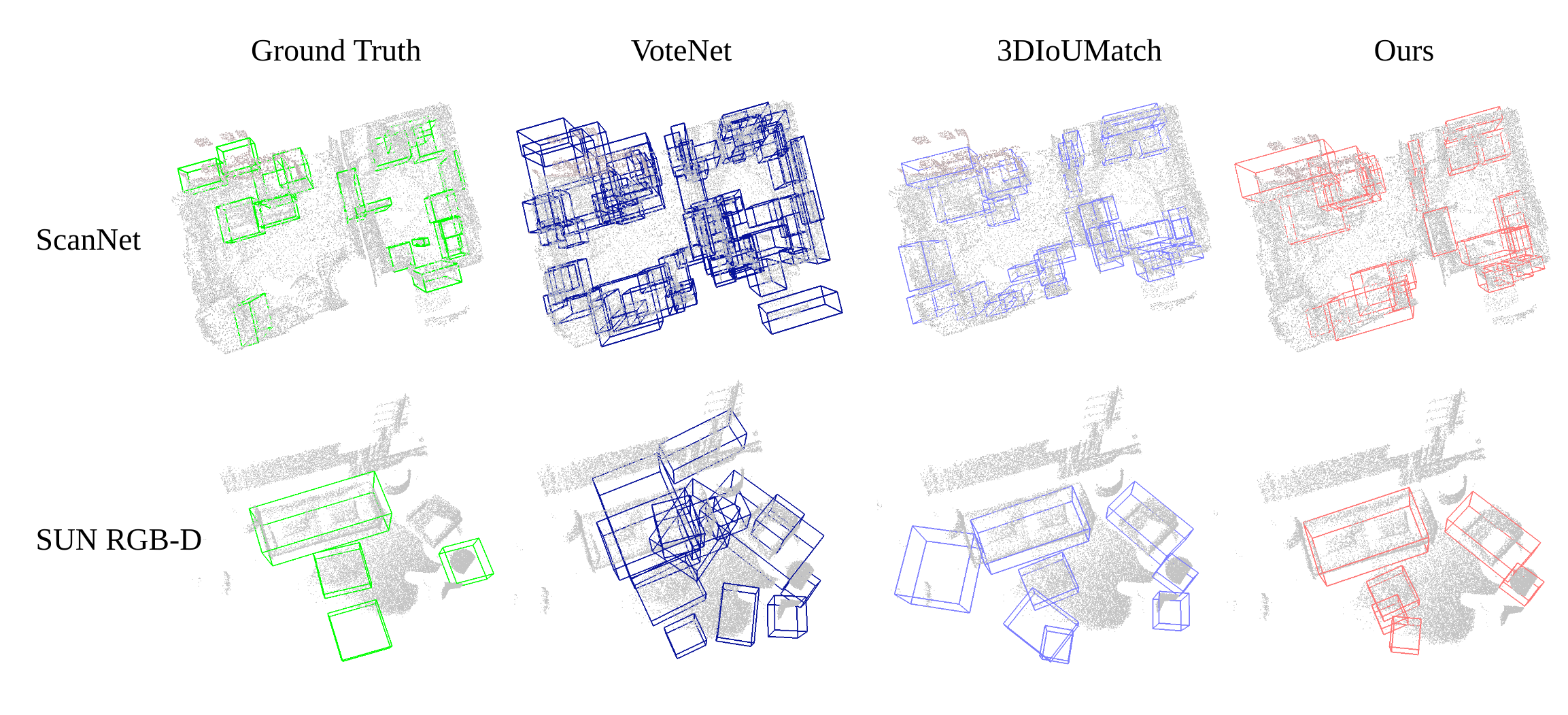}
\caption{Detection results comparison between the supervised baseline VoteNet~\cite{qi2019deep}, 3DIoUMatch~\cite{wang20213dioumatch}, and our method.} 
\label{fig:det-vis}
\end{figure*}

\begin{figure*}[ht]
    \centering
    \begin{subfigure}[b]{0.475\textwidth}  
        \centering 
        \includegraphics[width=\textwidth]{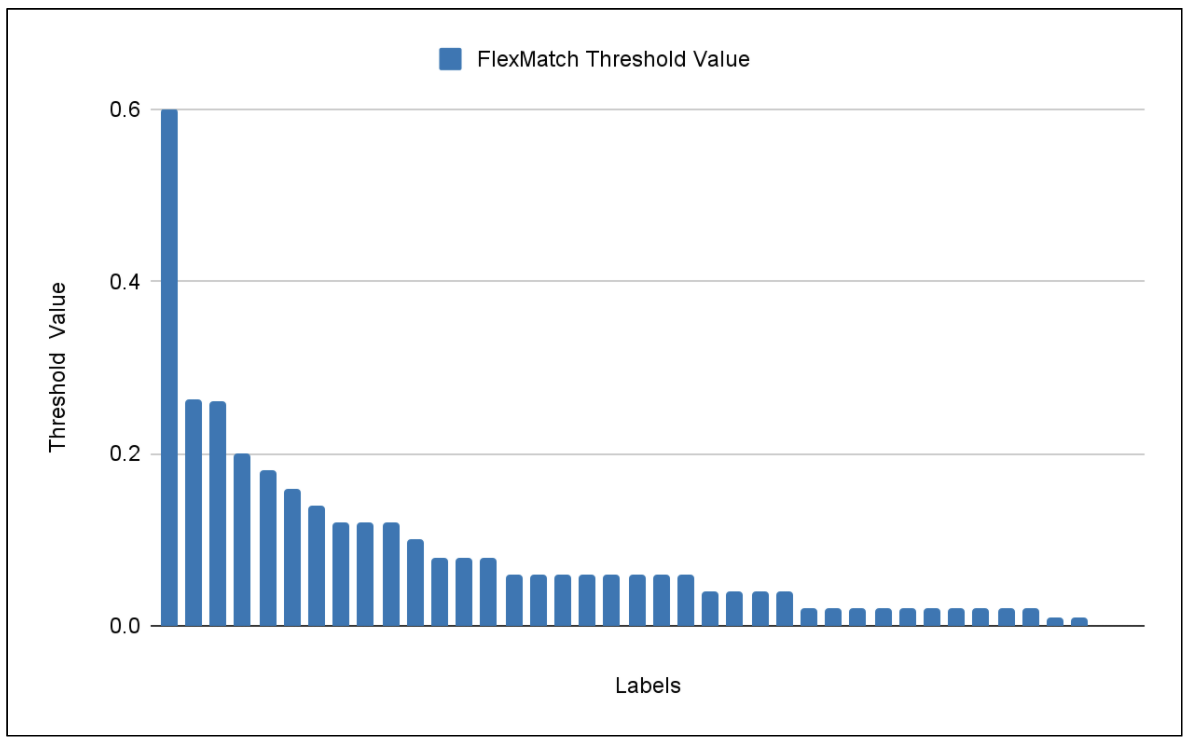}
        \caption[]%
        {{\small}}    
        \label{fig:flex-threshold}
    \end{subfigure}
    \hfill
    \begin{subfigure}[b]{0.475\textwidth}
        \centering
        \includegraphics[width=\textwidth]{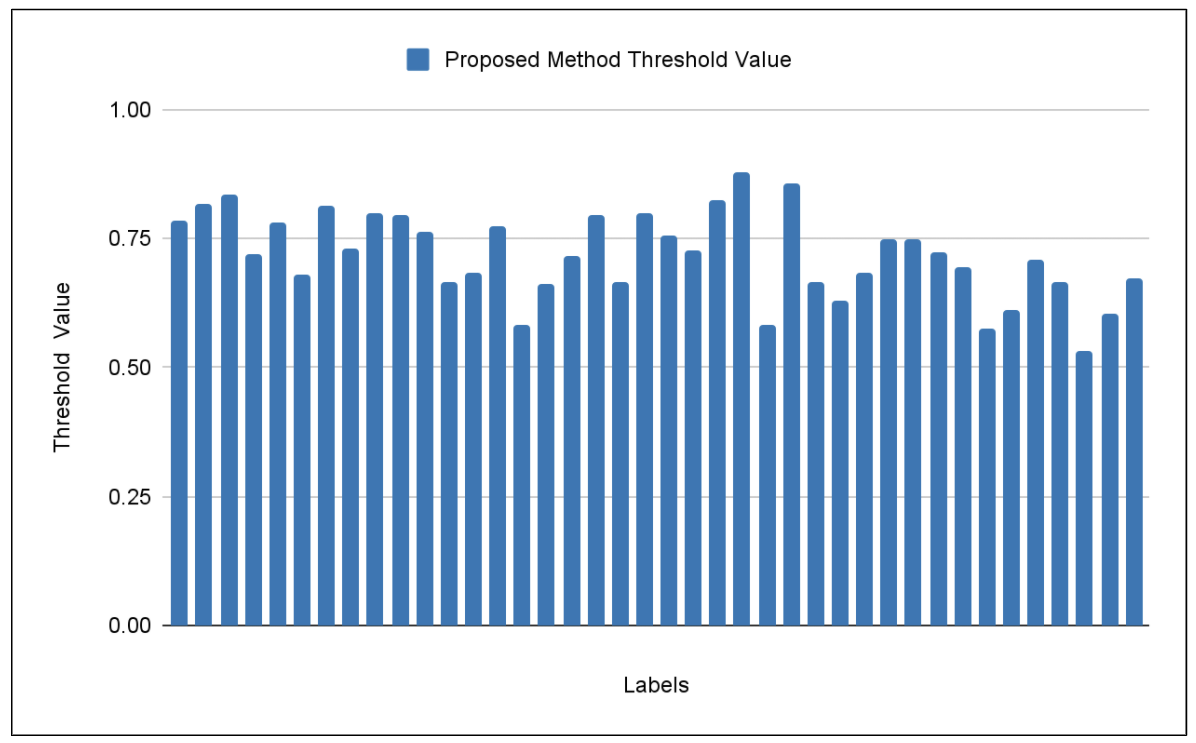}
        \caption[]%
        {{\small }}    
        \label{fig:our-threshold}
    \end{subfigure}
    \caption[ ]
    {\small (a) Thresholds of our method for each class in the last epoch under ModelNet40 dataset with $10$ percent labeled data. (b)  Thresholds of FlexMatch for each class in the last epoch. The FlexMatch leads to long-tail thresholds, which introduces much noise in pseudo labels and thus degrades the performance. Thresholds of our method is appropriate and balanced, which boosts the efficiency of utilizing unlabeled data.} 
    \label{fig:flexmatch-res}
\end{figure*}

\section{Comparison with Our Method and Dash}
Dash ~\cite{xu2021dash} proposed a dynamic threshold method based on cross entropy lose for all classes. However, the Latest SSL SOTA method FlexMatch ~\cite{zhang2021flexmatch} has demonstrated the benefit of class-level dynamic thresholding, which not only fully utilizes a large number of unlabeled data, but takes into account each class's learning status. Our method is inspired by FlexMatch and utilize class-level confidence to obtain dynamic threshold for each class. To further demonstrate the benefit of class-level threshold, we compare our method with Dash.  Table.~\ref{tab:imbalance_compa} shows that Dash only improves limited performance compared to baseline and our method achieves better performance than Dash in 3D tasks when only the dynamic threshold part is utilized.

\vspace{-2mm}
\begin{table}[hb]
\setlength{\belowcaptionskip}{-14pt}
\centering
\adjustbox{width=240pt,height=330pt,keepaspectratio=true}
{
	\begin{tabular}{c|c|c|c|c}

	    \hline
	     \multirowcell{2}{ } & 
	    \multicolumn{2}{c|}{ModelNet40 10\%} &
	    \multicolumn{2}{c}{SUN RGB-D 5\%} \\ \cline{2-5}
	   & Overall Acc & Mean Acc & mAP $@$0.25 & mAP $@$0.5
        \\
        \hline
        Baseline   & 85.5& 79.4  & 39.7 ± 0.9 & 20.6 ± 0.7   \\
        \hline
        Dash [38] + Baseline  & 85.9  &80.1& 40.5 ± 0.7 & 21.0 ± 0.6   \\
        \hline
        Confid-Threshold + Baseline  & 86.9  &81.7 & 42.0 ± 0.8 & 22.8 ± 0.5   \\
        \hline
        Ours + Baseline& \textbf{87.8}  & \textbf{82.5}  & \textbf{43.1 ± 0.6}  &  \textbf{24.2 ± 0.5}     \\
        \hline
	\end{tabular}
}
 \caption{\small Comparative studies with Dash.}
\label{tab:imbalance_compa}
\end{table}
\vspace{-3mm}

\begin{table*}[t!]
\centering
\captionsetup{font=footnotesize}
\adjustbox{width=390pt,height=800pt,keepaspectratio=true}
{

	\begin{tabular}{K{25mm}|K{15mm}|K{15mm}|K{15mm}|K{15mm}|K{15mm}|K{15mm}|K{15mm}|K{15mm}|K{15mm}|K{15mm}}

	    \hline
    	   & bathtub & bed & bookshelf  &chair &desk&  dresser   & nightstand & sofa & table  &toilet \\
    	\hline
    	\multicolumn{11}{c}{mAP@0.25} \\
    	\hline
    	VoteNet~\cite{qi2019deep}    &52.8  &28.8  &24.8  &48.1 &21.5 &
    	9.0    &0.3  &2.8 &0.8 &24.4  \\
    	3DIoUMatch~\cite{wang20213dioumatch}    &60.1  &\textbf{33.5}  &\textbf{36.5} & 55.9& 36.2& 6.7&0.3&
    	15.9    &3.6 &32.5    \\
    	Ours     &\textbf{62.8}  &30.3  &32.6  &\textbf{58.9}  &\textbf{54.2}&
    	\textbf{10.9}     &\textbf{0.8}  &\textbf{22.4}  &\textbf{3.8}  &\textbf{47.7}  \\
    	\hline
    	\multicolumn{11}{c}{mAP@0.5} \\
    	\hline
    	VoteNet~\cite{qi2019deep}    &16.7 &4.1  &7.1  &14.8  &3.1&
    	0.4    &0  &0.3& 0.2  & 4.4 \\
    	3DIoUMatch~\cite{wang20213dioumatch}    &20.0  &\textbf{9.7}  &18.8  &30.8 &\textbf{5.1}&
    	0.4    &0 &0.5  &0.8 &10.8  \\
    	Ours     &\textbf{25.3}  &8.7  &\textbf{19.3}  &\textbf{34.8} &2.2&
    	\textbf{1.1}     &0  &\textbf{9.4} &\textbf{1.2}  &\textbf{24.2} \\
    	
        \hline
	\end{tabular}
}
 \caption{Per-class performance comparison for the 3D object detection task with the state-of-the-art semi-supervised learning methods on the SUN RGB-D dataset with $2$ percent labeled data.}
\label{tab:detection-sun}
\end{table*}

\begin{table*}[t!]
\captionsetup{font=footnotesize}
\centering

\adjustbox{width=500pt,height=1200pt,keepaspectratio=true}
{

	\begin{tabular}{K{25mm}|K{15mm}|K{15mm}|K{15mm}|K{15mm}|K{15mm}|K{15mm}|K{15mm}|K{15mm}|K{15mm}|K{15mm}|K{15mm}|K{15mm}|K{15mm}|K{15mm}|K{15mm}|K{15mm}|K{15mm}|K{15mm}}

	    \hline
    	   & cab & bed & chair  &sofa &table&  door   & wind & bkshf & pic  &cntr &desk&curt&fridg&showr&toil&sink&bath&ofurn \\
    	\hline
    	\multicolumn{19}{c}{mAP@0.25} \\
    	\hline
    	VoteNet~\cite{qi2019deep}    &12.5  &65.9  &70.2  &69.2 &37.6 &
    	14.0    &8.5 &15.6 &0.5  &10.0 & 48.2    &16.8 &20.2 &20.6  &72.1&28.1&44.4&8.6   \\
    	3DIoUMatch~\cite{wang20213dioumatch}    &\textbf{27.0}  &\textbf{71.5}  &78.4  &72.3 &48.0 &
    	\textbf{22.9}    &\textbf{17.8}  &14.1 &1.6  &40.0 & 51.6    &25.0 &29.8 &43.6  &81.7&33.5&75.1&16.0  \\
    	Ours     &24.6  &69.6 & \textbf{79.0}  &\textbf{73.6}  &\textbf{49.1}&
    	18.5     &16.5  &\textbf{23.9}  &\textbf{3.5} &\textbf{41.7} &\textbf{62.5}  &\textbf{32.3} & \textbf{33.6} & \textbf{44.1} &  \textbf{93.2}& \textbf{33.8} &\textbf{78.6}  & \textbf{16.3}  \\

    	\hline
    	\multicolumn{19}{c}{mAP@0.5} \\
    	\hline
    	VoteNet~\cite{qi2019deep}    &0.1  &50.5  &34.3  &37.7  &16.1&
    	2.7    &1.2  &5.3 &0  & 1.2 & 13.1 &0.5 & 6.7&0 &49.0 &8.3 &27.5&0.9 \\
    	3DIoUMatch~\cite{wang20213dioumatch}    &3.2  &\textbf{57.0}  &56.1  &\textbf{53.2} &29.5&
    	\textbf{8.6}    &\textbf{4.9}  &4.7  &0 &2.1 &28.3 &3.1 & 15.7&7.3 &59.8 &6.2 &60.9 & 3.6 \\
    	Ours     &\textbf{4.1}  &56.6  &\textbf{56.9}  &50.9 &\textbf{30.7}&
    	6.2     &4.2  &\textbf{8.3} &0  &\textbf{3.2} &\textbf{31.6} & \textbf{11.1}& \textbf{23.2}&\textbf{7.6} & \textbf{60.0}&\textbf{9.0} &\textbf{62.9} &\textbf{4.6}\\
    	
        \hline
	\end{tabular}
}

 \caption{Per-class performance comparison for the 3D object detection task with the state-of-the-art semi-supervised learning methods on the ScanNet dataset with $5$ percent labeled data.}
\label{tab:detection-scan}
\end{table*}

\begin{table*}[t!]
\centering
\setlength{\belowcaptionskip}{-5pt}
\captionsetup{font=footnotesize}
\adjustbox{width=400pt,height=500pt,keepaspectratio=true}
{
	\begin{tabular}{K{20mm}|K{15mm}|K{15mm}|K{15mm}|K{15mm}|K{15mm}|K{15mm}|K{15mm}|K{15mm}}

	    \hline
    	   & bag & bin & box  &cabinet &chair&  desk   & display & door  \\
    	\hline
    	FixMatch~\cite{sohn2020fixmatch}    &2.4 &41.7  &\textbf{14.3}  &49.5 &83.1 &
    	\textbf{12.7}   &52.5&94.8    \\
    	FlexMatch~\cite{zhang2021flexmatch}    &1.1  &36.7  &0.5  &49.5 &85.9 &
    	3.3    &45.1  &\textbf{97.1}   \\
    	Ours    &\textbf{12.1}  &\textbf{42.7} & 12.8  &\textbf{63.4}  &\textbf{86.4}&
    	6.0     &\textbf{66.7}  &95.7    \\
    	\hline
    	& shelf  &table &bed&pillow&sink&sofa&toilet& \\
    	\hline
    	FixMatch~\cite{sohn2020fixmatch}    &38.6  &41.9  &46.4  &29.5 &41.7 &
    	39.5    &10.6 &   \\
    	FlexMatch~\cite{zhang2021flexmatch}    &35.7  &\textbf{61.1}  &\textbf{76.3}  &3.9 &51.2 &
    	50.5    &2.6  &     \\
    	Ours    &\textbf{41.5}  &55.9 & 60.9  &\textbf{34.8}  &\textbf{53.7}&
    	\textbf{87.6}    &\textbf{12.9}  &   \\
    	\hline
	\end{tabular}
}
\caption{Per-class performance comparison for the 3D object classification task with the state-of-the-art semi-supervised learning methods on the ScabObjectNN dataset with $2$ percent labeled data.}
\label{tab:classification-scan}
\end{table*}

\begin{table*}[t!]
\centering
\setlength{\belowcaptionskip}{-10pt}
\captionsetup{font=footnotesize}
\adjustbox{width=500pt,height=800pt,keepaspectratio=true}
{

	\begin{tabular}{K{20mm}|K{20mm}|K{20mm}|K{20mm}|K{20mm}|K{20mm}|K{20mm}|K{20mm}|K{20mm}|K{20mm}|K{20mm}}
	    \hline
    	   & airplane & bathtub & bed  &bench &bookshelf&  bottle   & bowl & car & chair  &cone  \\
    	\hline
    	FixMatch~\cite{sohn2020fixmatch}    &100  &\textbf{78}  &\textbf{99}  &65 &98 &95
    	&85   &94 &98 &\textbf{100}     \\
    	FlexMatch~\cite{zhang2021flexmatch}    &100  &62  &99  &60 &97 &
    	\textbf{97}    &\textbf{90} &\textbf{99} &\textbf{100} &100   \\
    	Ours    &\textbf{100}  &76  &98  &\textbf{65} &\textbf{98} &
    	96    &86 &98 &98  &95   \\
    	\hline
    	&cup& curtain & desk  &door &dresser&  flower pot   & glass box & guitar & keyboard  &lamp  \\
    	\hline
    	FixMatch~\cite{sohn2020fixmatch}    &30  &55  &\textbf{81}  &80 &\textbf{88} &20
    	    &91 &99 &95  &65   \\
    	FlexMatch~\cite{zhang2021flexmatch}    &40 &50 &73  &85 &76 &
    	25    &91 &100 &100 &65   \\
    	Ours    &\textbf{50}  &\textbf{70} &74  &\textbf{90} &83 &
    	\textbf{30}    &\textbf{92} &\textbf{100} &\textbf{100}  &\textbf{85}    \\
    	\hline
    	   & laptop & mantel & monitor  &night stand &person&  piano   & plant & radio & range hood  &sink  \\
    	\hline
    	FixMatch~\cite{sohn2020fixmatch}  & 100 &92  &95  &44  &65 &85 &
    	\textbf{85}    &50 &85 &60      \\
    	FlexMatch~\cite{zhang2021flexmatch}    &100  &93  &99  &47 &75 &
    	93    &80 &60 &72  &\textbf{80}   \\
    	Ours    &\textbf{100}  &\textbf{93} &\textbf{99}  &\textbf{53} &\textbf{75} &
    	\textbf{95}    &78 &\textbf{60} &\textbf{86}  &75    \\
    	\hline
    	   & sofa & stairs & stool  &table &tent&  toilet   & tv stand & vase & wardrobe  &xbox  \\
    	\hline
    	FixMatch~\cite{sohn2020fixmatch}  &  \textbf{100}&65  &\textbf{80}  &90  &95 &99 &
    	61   &81 &35 &75      \\
    	FlexMatch~\cite{zhang2021flexmatch}    &98  &90  &55  &\textbf{91} &95 &
    	99    &\textbf{85} &\textbf{88} &25  &65  \\
    	Ours    &99  &\textbf{90} &70  &90 &\textbf{95} &
    	\textbf{99}    &82 &85 &\textbf{45}  &\textbf{77}  \\
    	\hline
    	
	\end{tabular}
}
\caption{Per-Class performance comparison for the 3D object classification task with the state-of-the-art semi-supervised learning methods on the ModelNet40 dataset with $10$ percent labeled data. In high learning status classes, performances of FixMatch, FlexMatch, and our method are similar. In low learning status classes, our method outperforms the FixMatch and FlexMatch by a large margin.}
\label{tab:classification-modelnet}
\end{table*}

\section{Comparison with Our Method and FlexMatch}

To utilize more unlabeled data at the early stage of the training, the Flexmatch~\cite{zhang2021flexmatch} proposes a threshold warm-up strategy, which decreases the threshold according to the number of unused unlabeled data. However, due to the high learning difficulty of 3D data, a large number of unlabeled data remains unused during the training, which decreases the dynamic threshold of each class when FlexMatch is used. Furthermore, FlexMatch adjusts the threshold of each class according to the pseudo-labels numbers for each class. It works well for class-balanced dataset, but in commonly used 3D data~\cite{dai2017scannet,uy-scanobjectnn-iccv19,song2015sun,wu20153d}, the numbers of labeled data in each class is long-tail and thus the numbers of high confidence pseudo-labeled data is also tend to be long-tail. For example, in ModelNet40, in ModelNet40, the label numbers of airplane and bowl are 563 and 59 separately. Even if the dynamic threshold filters half pseudo-labels of airplane and utilizes all pseudo-labels of bowls, the airplane’s selected unlabeled data numbers are at least four times larger than the bowl’s selected unlabeled data numbers. This makes the threshold value of airplane much larger than the bowl in FlexMatch. Hence, as demonstrated in Fig.~\ref{fig:flex-threshold}, utilizing FlexMatch generates low and significantly variant (long-tail) thresholds, which introduces much noise, especially for those minority classes, and thus achieves unsatisfied performances in 3D tasks. Unlike FlexMatch adjusting thresholds based on pseudo-label numbers, our method utilizes class-level confidence to adjust dynamic thresholds. As shown in the Fig.~\ref{fig:our-threshold} our method produces appropriate and balanced dynamic thresholds, even when dataset is imbalanced. Hence, our method boosts the efficiency of utilizing unlabeled data without introducing much noise and have more generality than FlexMatch.

\section{Per-class Accuracy for Classification}

To understand the performance of our method on each class in the 3D SSL classification task, we report per-class accuracy on ScanObjectNN with $2$ percent labeled data and ModelNet40 with $10$ percent labeled data, respectively. The Table~\ref{tab:classification-scan} indicates that in most classes, our method has better performance than FixMatch~\cite{sohn2020fixmatch} and FlexMatch~\cite{zhang2021flexmatch} . Moreover, In low learning status classes, the FlexMatch even degrades the performance of FixMatch due to the noise brought by low thresholds. Table~\ref{tab:classification-modelnet} shows that in high learning status classes, performances of FixMatch, FlexMatch, and our method are similar. In low learning status classes, our method outperforms the FixMatch and FlexMatch by a large margin. This demonstrates our method's capability in improving low learning status classes and re-balance network's learning statuses.

\section{Per-class Accuracy for Detection}
\vspace{-3pt}
To understand the performance of our method on each class in the 3D SSL detection task, we report per-class accuracy on SUN RGB-D with $2$ percent labeled data and ScanNet with $5$ percent labeled data, respectively. From Table~\ref{tab:detection-sun} and~\ref{tab:detection-scan} we can find that in  most  classes, our method has better performances than 3DIoUMatch~\cite{wang20213dioumatch} and backbone VoteNet~\cite{qi2019deep}.

\begin{figure*}[t!]
    \centering
    \begin{subfigure}[b]{0.3\textwidth}
        \centering
\includegraphics[width=\textwidth]{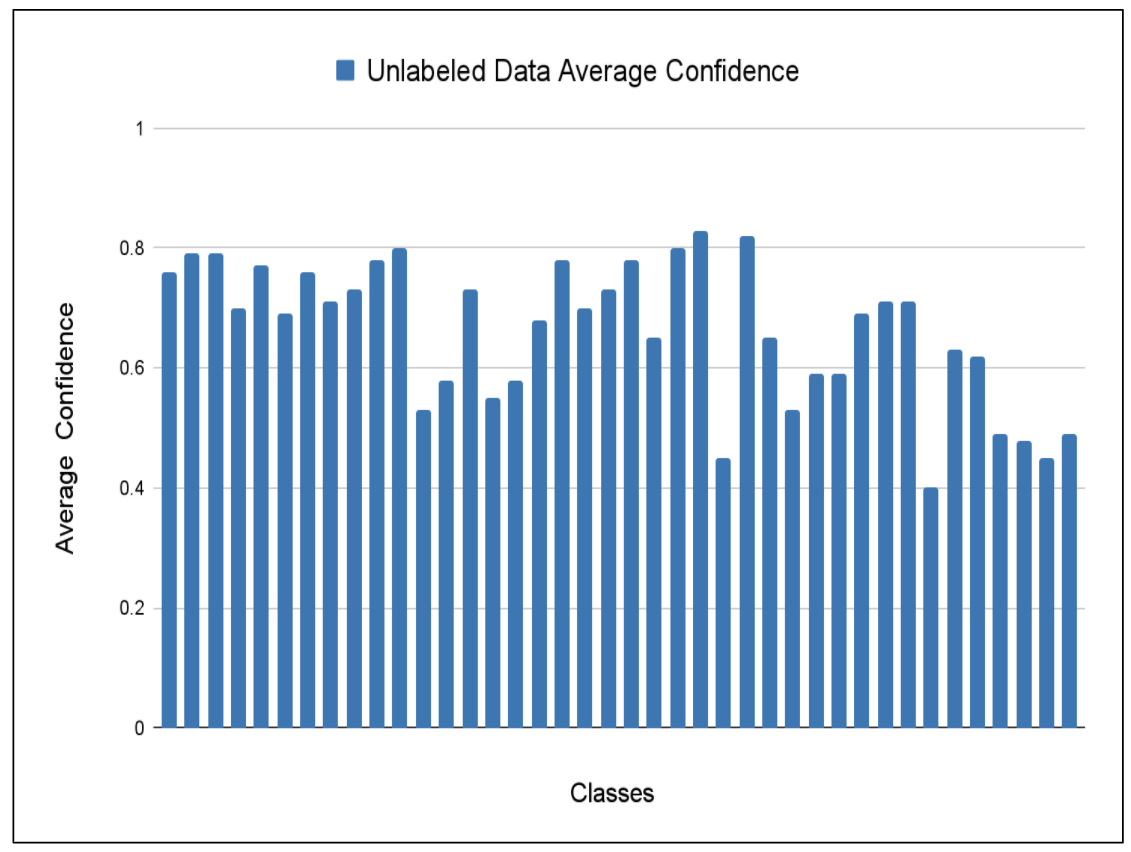}
        \caption[]%
        {{\small FixMatch average confidence}}    
        \label{fig:modelnet_fixmatch_aveconfid}
    \end{subfigure}
    \begin{subfigure}[b]{0.3\textwidth}   
        \centering 
\includegraphics[width=\textwidth]{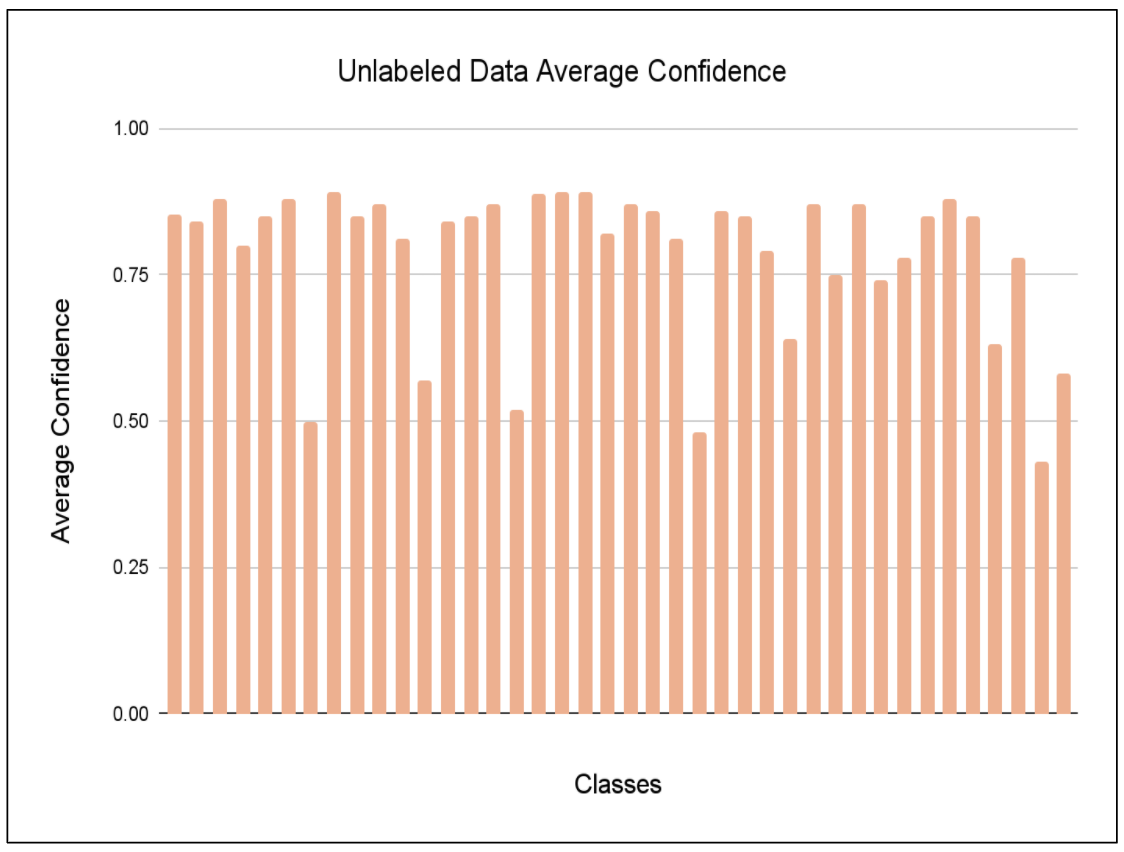}
        \caption[]%
        {{\small FlexMatch average confidence.}}    
        \label{fig:modelnet_flexmatch_aveconfid}
    \end{subfigure}
    \begin{subfigure}[b]{0.3\textwidth}   
        \centering 
\includegraphics[width=\textwidth]{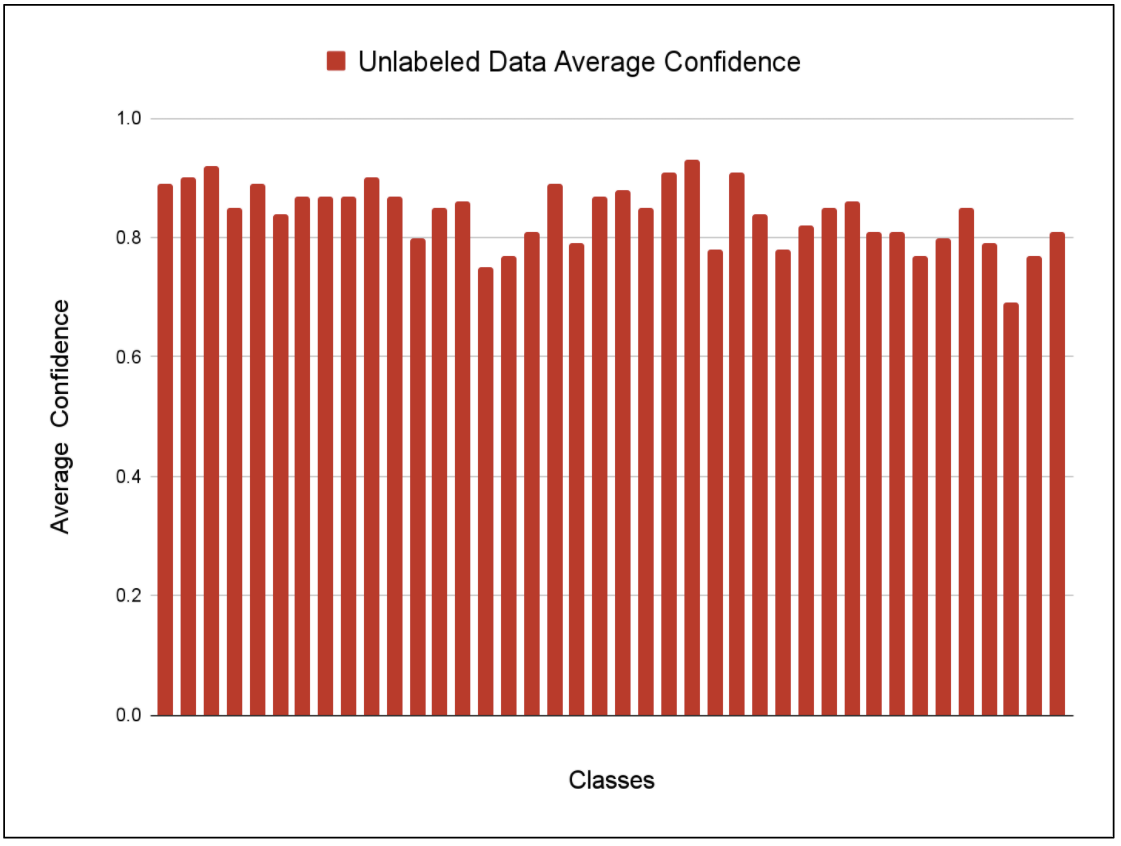}
        \caption[]%
        {{\small Our method average confidence.}}    
        \label{fig:modelnet_our_aveconfid}
    \end{subfigure}
    \vspace{-5pt}

    \caption[ The average and standard deviation of critical parameters ]
    {\small The comparison of unlabeled data class-level confidence between FixMatch and our method in ModelNet40 datasets trained with $10$ percent labeled data. The result demonstrates that our method not only improves the learning status of each class but makes the learning status more balanced.} 
    \label{fig:confidence comparison}
    \vspace{-15pt}
\end{figure*}

\section{Result Analysis and Discussion} 

To take a deeper look at how our model improves the performance, we conduct analysis about the results of our model and compare with FixMatch, which uses a fixed threshold $0.90$ and FlexMatch. We calculated the average class-level confidence of FixMatch, FlexMatch, and our method on ModelNet40 datasets.


Fig.~\ref{fig:modelnet_fixmatch_aveconfid}  shows that the class-level confidence of FixMatch model is imbalanced and relatively lower. This is probably caused by the fixed threshold since it does not consider the learning difficulty and status of different classes. The Fig.~\ref{fig:modelnet_flexmatch_aveconfid} shows that although the FlexMatch improve confidence of some classes, the class-level confidence still remains imbalanced. This is because FlexMatch is not designed for data-imbalanced dataset and thus cannot re-balance the learning status.  The class-level confidence of our proposed method is shown in the Fig.~\ref{fig:modelnet_our_aveconfid}. Benefiting from the dynamic threshold and re-sampling strategy, our proposed method not only improves classes' average confidence but also makes learning status balanced compared to the FixMatch and FlexMatch. This analysis confirms the advantage of using the dynamic threshold for each class.

{\small
\bibliographystyle{ieee_fullname}
\bibliography{egbib}
}